\documentclass[10pt, conference, letterpaper]{IEEEtran}
\IEEEoverridecommandlockouts

\usepackage{cite}
\usepackage{amsmath,amssymb,amsfonts}
\usepackage{algorithmic}
\usepackage{algorithm}
\usepackage{graphicx}
\usepackage{textcomp}
\usepackage{xcolor}
\def\BibTeX{{\rm B\kern-.05em{\sc i\kern-.025em b}\kern-.08em
    T\kern-.1667em\lower.7ex\hbox{E}\kern-.125emX}}
\usepackage{setspace}
\usepackage{enumitem}
\usepackage{cleveref}
\usepackage{subfigure}
\usepackage{caption}
\usepackage{multirow}
\usepackage{tabularx}
\usepackage{booktabs}

\newcounter{tag}
\crefname{section}{§}{§§}
\Crefname{section}{§}{§§}

\begin{document}

\title{CODA: A Continuous Online Evolve Framework for Deploying HAR Sensing Systems
}

\author{
    \IEEEauthorblockN{Minghui Qiu$^{1}$, Jun Chen$^{1}$, Lin Chen$^{1}$, Shuxin Zhong$^{1*}$, Yandao Huang$^{1, 2}$, Lu Wang$^{3}$, Kaishun Wu$^{1}$\thanks{$^{*}$ is the corresponding author (shuxinzhong@hkust-gz.edu.cn).}}
    
    \IEEEauthorblockA{$^1$The Hong Kong University of Science and Technology (Guangzhou), Guangzhou, China}
    \IEEEauthorblockA{$^2$The Hong Kong University of Science and Technology, Clear Water Bay, Hong Kong}
    \IEEEauthorblockA{$^3$Shenzhen University, Shenzhen, China}
    
}
  

\maketitle


\begin{abstract}


In always-on HAR deployments, model accuracy erodes silently as domain shift accumulates over time.
Addressing this challenge requires moving beyond one-off updates toward instance-driven adaptation from streaming data.
However, continuous adaptation exposes a fundamental tension: 
systems must selectively learn from informative instances while actively forgetting obsolete ones under long-term, non-stationary drift.
To address them,
we propose \textbf{CODA}, a continuous online adaptation framework for mobile sensing. 
CODA introduces two synergistic components: 
(i) \textit{Cache-based Selective Assimilation}, which prioritizes informative instances likely to enhance system performance under sparse supervision, 
and (ii) an \textit{Adaptive Temporal Retention Strategy}, which enables the system to gradually forget obsolete instances as sensing conditions evolve. 
By treating adaptation as a principled cache evolution rather than parameter-heavy retraining, CODA maintains high accuracy without model reconfiguration. 
We conduct extensive evaluations on four heterogeneous datasets spanning phone, watch, and multi-sensor configurations. 
Results demonstrate that CODA consistently outperforms one-off adaptation under non-stationary drift, remains robust against imperfect feedback, and incurs negligible on-device latency.

\end{abstract}


\section{Introduction}
\begin{figure}[htb]
  \centering
    \includegraphics[width=.8\linewidth]{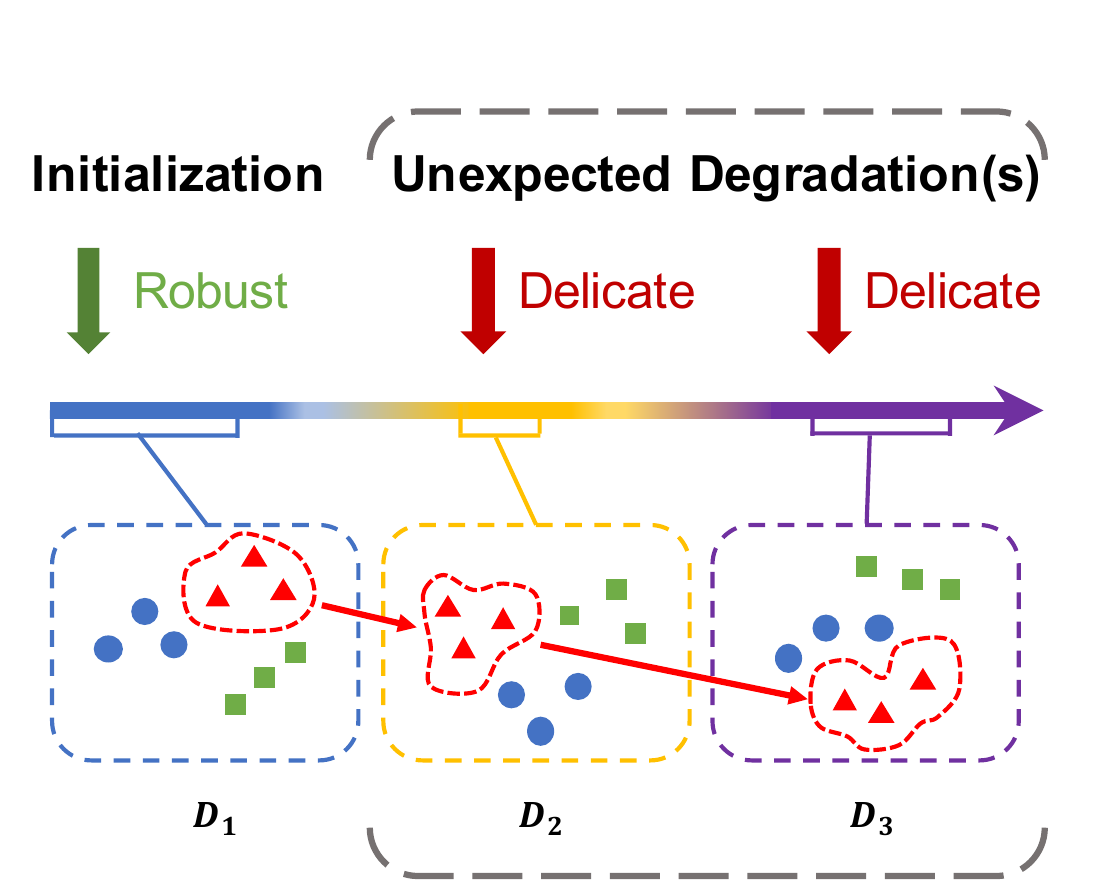}
  \caption{Unexpected degradations of the system at deployment time (in durations and directions). }
  \label{fig:IDea_illus}
  \vspace{-1em}
\end{figure}

Human Activity Recognition (HAR) systems are now routinely embedded in long-term, real-world applications, 
ranging from healthcare monitoring\cite{chen2019taprint, pan2025cgmm} and assisted living\cite{teng2023pdges}, to fitness tracking\cite{liu2024imove}, and smart homes\cite{kim2024iris,thukral2025layout}.
Unlike controlled laboratory settings, these systems are expected to run continuously in the background, while users change, behaviors evolve, devices shift on the body, and surrounding environments fluctuate over time (as illustrated in Fig. \ref{fig:IDea_illus}).
In such always-on deployments, empirical studies~\cite{chen2021deep} consistently report that model performance degrades over time, even without explicit failures or abrupt domain transitions.
As these degradations accumulate silently during prolonged operation~\cite{yoon2025selfreplay}, understanding and mitigating this form of long-term, implicit domain shift remains a pressing challenge for building robust, real-world HAR systems~\cite{hong2024crosshar}.


Existing approaches to mitigating domain shift in HAR primarily fall into supervised and self-supervised paradigms.
Supervised methods include domain generalization~\cite{lu2022semantic, qian2021latent, qin2022domain}, which seeks to learn domain-invariant representations, and domain adaptation, which exploits access to labeled target-domain data to specialize models to a specific deployment environment~\cite{chang2020systematic, gong_metasense_2019, wang2018stratified, zhou2020xhar}.
Self-supervised approaches further incorporate meta-learning techniques to improve adaptation efficiency by learning transferable initialization or update rules across tasks or domains\cite{yoon2025selfreplay}.
Despite their effectiveness under controlled settings, these methods largely share a common assumption: 
adaptation is treated as a one-time process, performed either before deployment or at a clearly defined adaptation stage.
Such assumptions break down in real-world mobile sensing deployments, where domain shifts emerge continuously and unpredictably as user behaviors evolve beyond the control of system designers or service providers, 
causing one-off strategies to fall short of sustaining the long-term reliability required by always-on systems~\cite{ramasamy2018recent, hong2024crosshar}.

This gap points to a fundamental shift in how adaptation should be approached.
In continuous operating system, adaptation cannot rely on sporadic model updates; 
instead, it must be driven by the data as it arrives during normal operation.
Each incoming sensing instance implicitly captures the system's current operating conditions, including user behavior, device configuration, and environmental context~\cite{hong2024crosshar}.
Importantly, domain drift does not erase structure~\cite{gong_metasense_2019}.
Even after drift occurs, instances corresponding to the same activity often remain locally structured in the feature space, offering a natural yet underutilized opportunity for instance-driven adaptation. 

However, exploiting this opportunity is far from trivial and exposes two fundamental challenges in continuous adaptation.
First, \textbf{not all instances contribute equally to adaptation (C1)}.
While informative instances can refine decision boundaries, noisy or unrepresentative instances may instead introdues irreversible bias if incorporated indiscriminately.
Second, under long-term, non-stationary shift, \textbf{previously useful instances may gradually become obsolete (C2)},
and naively accumulating historical data risks anchoring the model to outdated behaviors.
Therefore, these challenges reveal a core tension in continuous adaptation:
systems must selectively learn from the present while actively forgetting the past.

To resolve this tension, we propose CODA, 
a sustainable evolution framework for always-on mobile sensing deployments that 
reconceptualizes adaptation as an instance-centric evolutionary process, enabling models evolve continuously as runtime data arrive.
CODA consists of two key components, each designed to address one fundamental challenge.
\textbf{To address the heterogeneity in instance importance (C1)}, 
CODA introduces \textit{Cache-based Selective Assimilation}, 
a mechanism that regulates how and to what extent runtime instances are premitted to influence model evolution.
By leveraging active learning signals to estimate instance importance under sparse and imperfect supervision,
CODA selectively assimilates informative instances while attenuating the impact of noisy or misleading instances.
\textbf{To cope with long-term and non-stationary drift (C2)}, CODA further proposes \textit{Adaptive Temporal Retention Strategy} that explicitly models the temporal relevance of historical instances.
Rather than indiscriminately retaining past data, CODA maintains a bounded instance memory and continuously reweights retained instances over time, allowing obsolete or less relevant instances to be progressively forgotten as sensing conditions evolve.
Together, these designs enable CODA to achieve stable yet plastic adaptation—preserving useful historical knowledge while remaining responsive to emerging behaviors—thereby sustaining robust long-term performance in real-world mobile sensing systems.
The contributions are listed as follows:
\begin{itemize} [leftmargin=*]
\item We present CODA, the first continuous online adaptation framework that redefines adaptation as a sustained evolutionary process, enabling robust learning under long-term, non-stationary domain drift in always-on deployments, without relying on explicit adaptation phases or domain boundary signals.

\item CODA designs two key components: 
i) \textit{Cache-based Selective Assimilation}, which regulates how and to what extent runtime instances influence model evolution by prioritizing informative samples while suppressing noisy or misleading ones under sparse and imperfect feedback;
and ii) \textit{Adaptive Temporal Retention Strategy}, which explicitly models the temporal relevance of historical data, enabling obsolete instances to be progressively forgotten as sensing conditions evolve. 

\item  Through extensive evaluations conducted on 2 public and 2 self-collected HAR datasets, we demonstrate that CODA consistently outperforms one-off adaptation under continuous domain drift, remains robust with sparse and imperfect feedback, and incurs only negligible latency overhead in real-world mobile deployments.

\end{itemize}

\section{Related Work}\label{section:background}


\subsection{HAR in Always-On Mobile Sensing Systems}
Human activity recognition (HAR) has become a core sensing primitive in a wide range of always-on applications, including healthcare monitoring\cite{chen2019taprint, pan2025cgmm}, assisted living\cite{kim2024iris,thukral2025layout, teng2023pdges}, and fall detection\cite{waffle, wang2016wifall}.
Unlike controlled experimental settings, real-world HAR systems operate continuously under evolving users, device configurations, and environmental conditions~\cite{chen2021deep},
inevitably inducing distribution shifts between training and deployment data~\cite{yoon2025selfreplay,stisen_smart_2015, ustev2013user}.
Thus, models that perform well in controlled settings often suffer substantial performance degradation after deployment~\cite{ min_early_2019,yang_toward_2009,ustev2013user}.


\subsection{Domain Adaptation Techniques}
To mitigate performance degradation caused by domain shift, numerous domain adaptation and generalization methods have been proposed.
Broadly, existing approaches fall into three categories: data-level operations, representation learning, and learning strategies.
Data-level methods improve generalization by manipulating training data to better cover potential target domains \cite{lee2025domclp}.
These methods assume that domain discrepancies can be alleviated by reshaping or enriching the training data distribution.
Representation learning approaches focus on learning domain-invariant features through feature alignment or invariant network designs\cite{qin2022domain, cheng2024disentangled, yu2024clipceil}.
Learning strategy–based methods, including meta-learning and ensemble learning, leverage prior knowledge to improve adaptation efficiency \cite{gong_metasense_2019, yoon2025selfreplay}.
Recent HAR studies further integrate meta-learning with self-supervised objectives, such as CrossHAR and ContrastSense, to enhance cross-dataset generalization under limited labels\cite{lu2022semantic, qian2021latent, qin2022domain, hong2024crosshar, dai2024contrastsense}.
\textit{Despite their effectiveness, 
these methods largely rely on an episodic adaptation paradigm, in which adaptation is performed as a one-off process under the assumption of a well-defined target domain or a finite adaptation stage.
Such assumptions fundamentally break down in real-world mobile sensing deployments, where domain drift emerges continuously and unpredictably over time.}

\section{Methodology} \label{section:SysDetails}



\begin{figure}[t]

  \centering

    \includegraphics[width=0.9\linewidth]{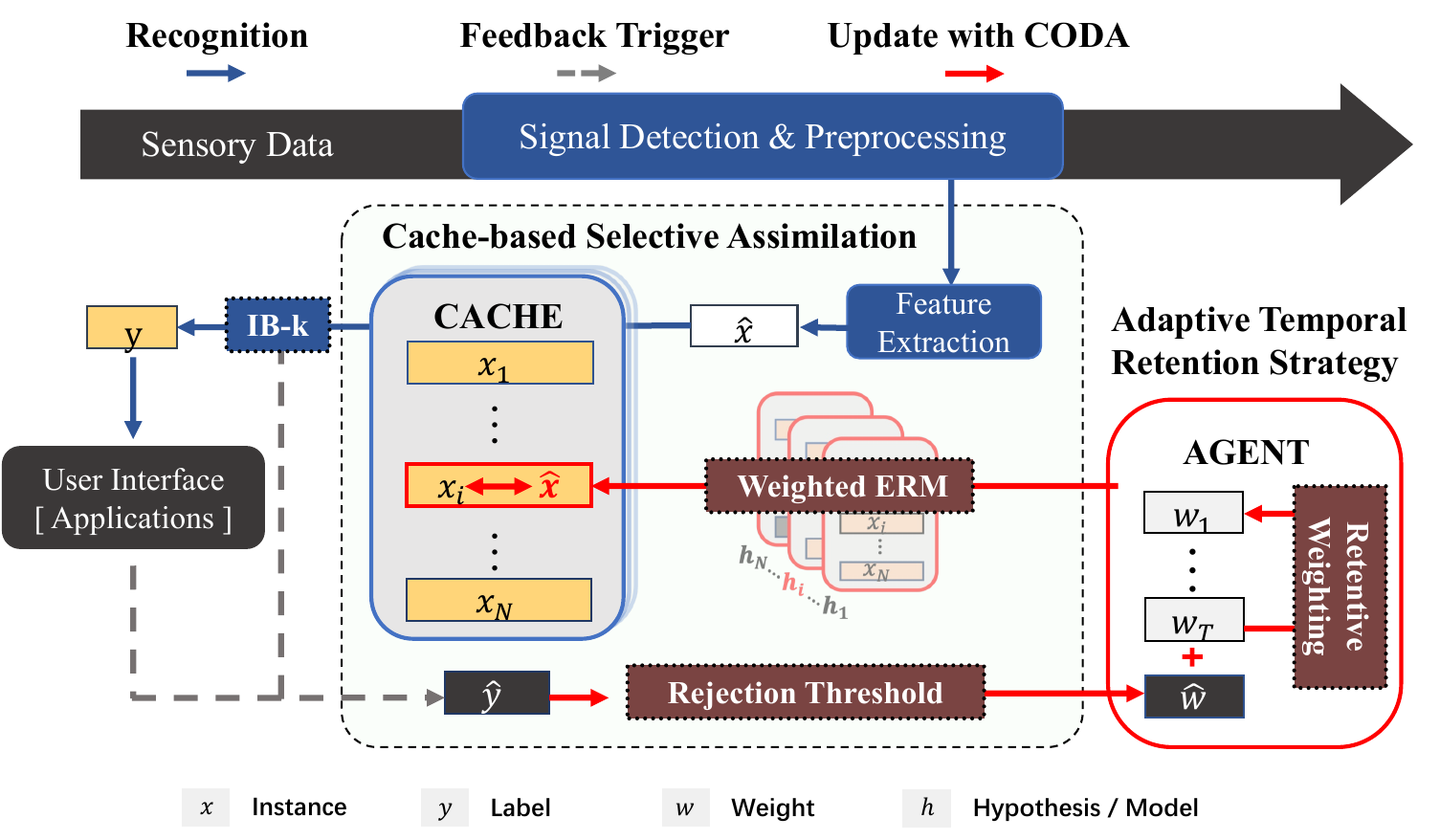}
  \caption{Adaptation pipeline in CODA (at time $T$).}
  \label{fig:overview}
  \vspace{-1em}
\end{figure}

To address the domain shift that accumulates during long-term operation, we propose CODA, a continuous, instance-level, deployment-time adaptation framework for human-centric sensing. 
As illustrated in Fig.~\ref{fig:overview}, CODA consists of two main mechanisms:
\textit{Cache-based Selective Assimilation} and 
\textit{Adaptive Temporal Retention Strategy}.

\begin{figure}[t]
\centering
\scalebox{0.88}{
\begin{minipage}{\linewidth}

\begin{algorithm}[H]
\setstretch{1.1}
\caption{Refined Importance-Weighted Active Learning}
\label{alg:IWAL}
\begin{algorithmic}[1]
\setstretch{1.5}
\STATE Initialize cache $S_0 = \emptyset$
\FOR{$t = 1,2,3,\dots$}
    \STATE Receive instance $(x_t, \hat{y}_t)$
    \STATE $(p_t, H_t) \leftarrow \textbf{Rejection-Threshold}(x_t, \hat{y}_t, S_{t-1})$
    \STATE Draw $Q_t \sim \text{Bernoulli}(p_t)$
    \IF{$Q_t = 1$}
        \STATE $S_t \leftarrow S_{t-1} \cup \{(x_t, \hat{y}_t, w_t, t)\}$
    \ELSE
        \STATE $S_t \leftarrow S_{t-1}$
    \ENDIF
    \STATE $h_t \leftarrow \arg\min\limits_{h \in H_t} \mathcal{L}_t(h, S_t)$
    \hfill $\triangleright$ \textbf{Weighted ERM}
\ENDFOR
\end{algorithmic}
\end{algorithm}

\vspace{-1em}

\begin{algorithm}[H]
\setstretch{1.1}
\caption{Rejection-Threshold }
\label{alg:Rejection}
\begin{algorithmic}[1]
\setstretch{1.5}
\REQUIRE Instance $x_t$, feedback $y_t$, cache $S_{t-1}$
\ENSURE Importance probability $p_t$, pruned hypothesis space $H_t$
\STATE Initialize hypothesis space $H_{t-1}$
\STATE $\mathcal{L}_{t-1}^{*} \leftarrow 
\min\limits_{h \in H_{t-1}} \mathcal{L}_{t-1}(h, S_{t-1})$
\STATE $H_t \leftarrow \left\{
h \in H_{t-1} :
\mathcal{L}_{t-1}(h, S_{t-1})
\le \mathcal{L}_{t-1}^{*} + \Delta_{t-1}
\right\}$
\STATE $p_t \leftarrow 
\max\limits_{f,g \in H_t,\, y \in Y}
\left[ \ell(f(x_t), y) - \ell(g(x_t), y) \right]$
\STATE \textbf{return} $p_t, H_t$
\end{algorithmic}
\end{algorithm}

\end{minipage}
}
\label{fig:algorithms}
\end{figure}

\subsection{Cache-based Selective Assimilation}

\begin{figure}[t]
    \centering 
    \setlength{\abovecaptionskip}{0.1cm}
    \setlength{\belowcaptionskip}{0.1cm}
    
    \begin{minipage}[t]{0.4\linewidth}
        \centering 
        \includegraphics[width=.95\linewidth]{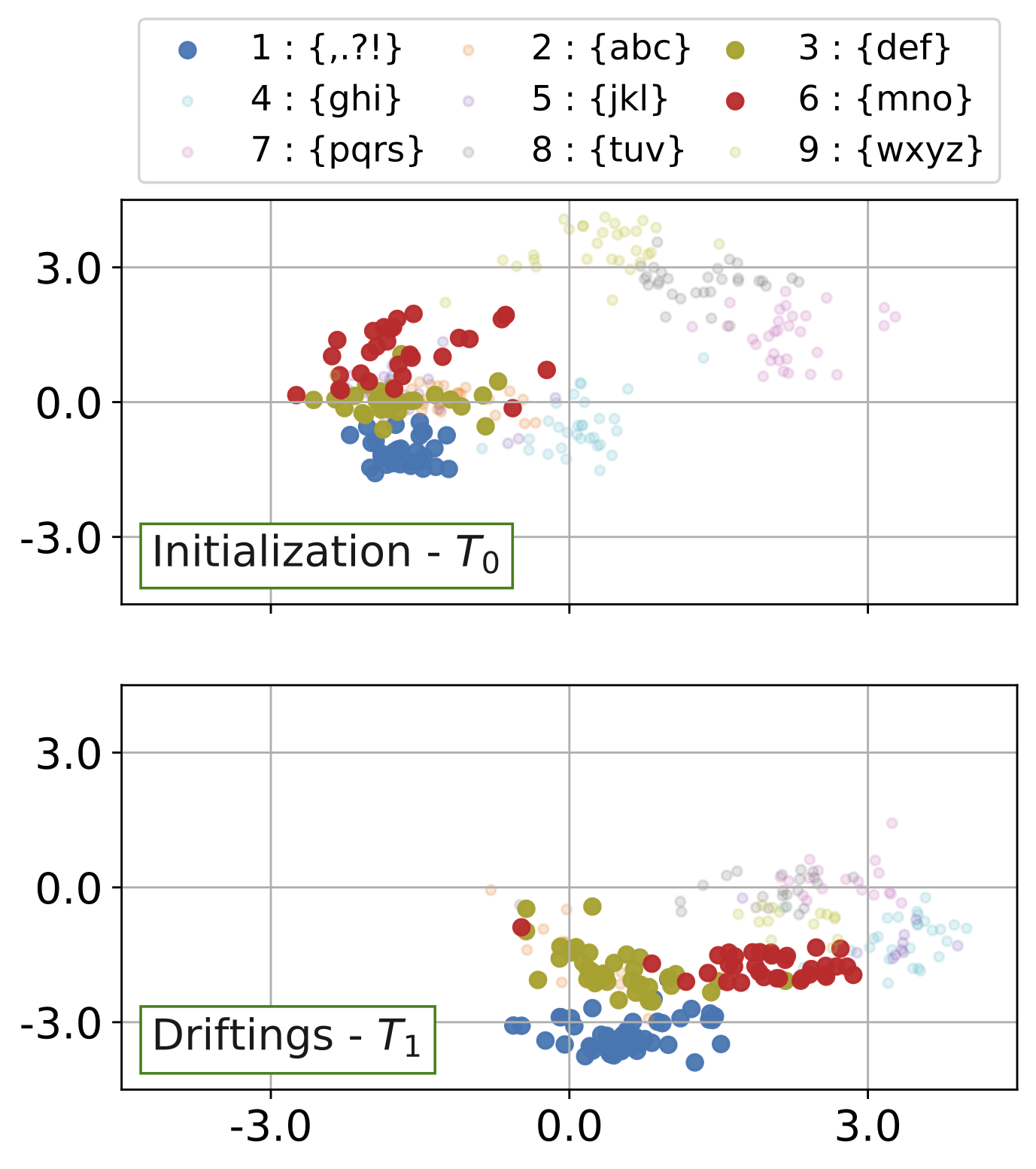}
        \caption*{\small (a) TAPRINT.}
    \end{minipage}
    \begin{minipage}[t]{0.4\linewidth}
        \centering 
        \includegraphics[width=.95\linewidth]{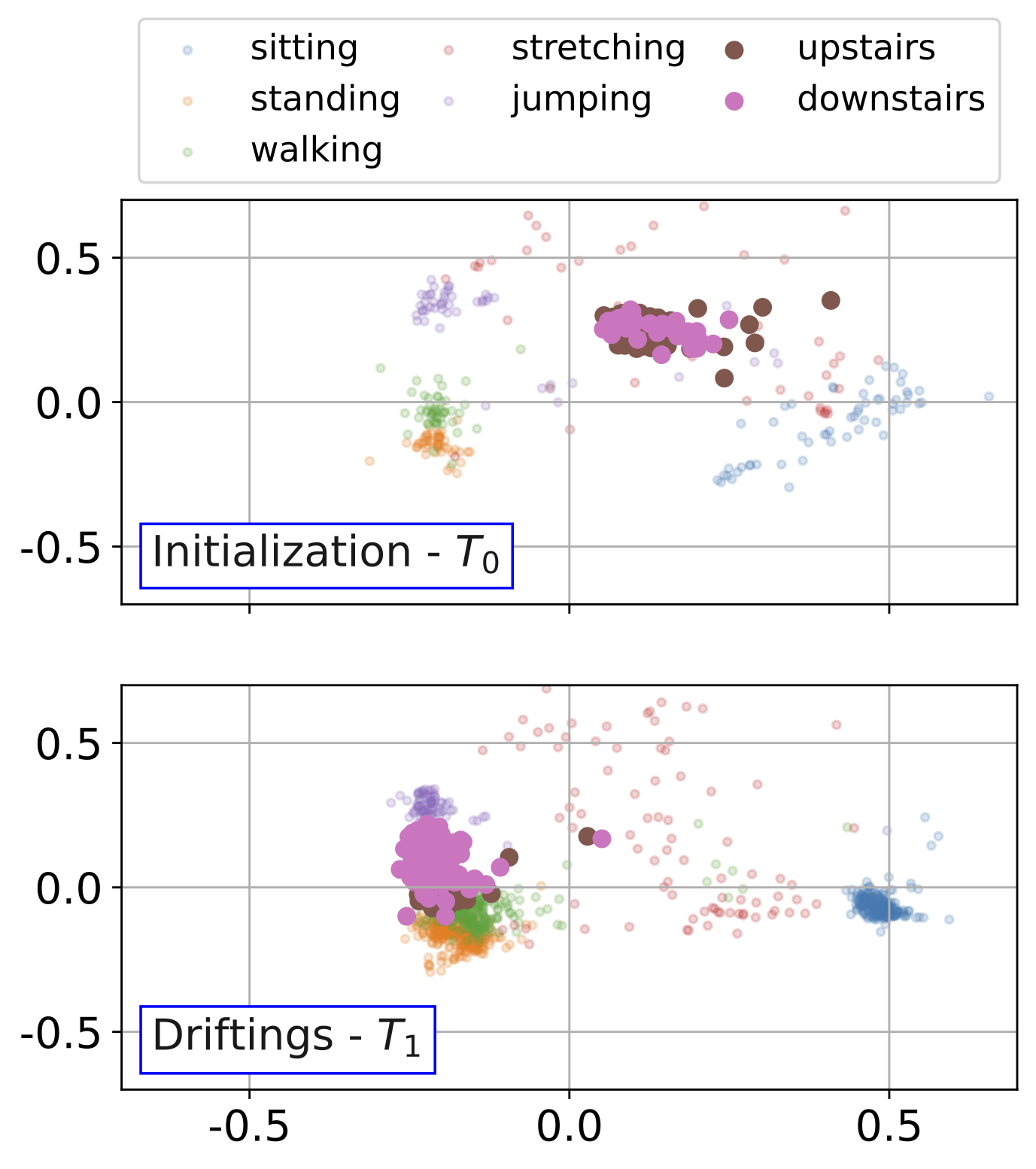} 
        \caption*{\small (b) WHAR.}
    \end{minipage}
    
    \caption{We observe the driftings at time $T_1$ comparing to the initialization phase at time $T_0$ with \underline{hand-tune} features.}
      \vspace{-1em}
    \label{fig:motivate_1}
\end{figure}

\begin{figure}[t]
    \centering 
    \setlength{\abovecaptionskip}{0.1cm}
    \setlength{\belowcaptionskip}{0.1cm}
    
    \begin{minipage}[t]{0.4\linewidth}
        \centering 
        \includegraphics[width=0.95\linewidth]{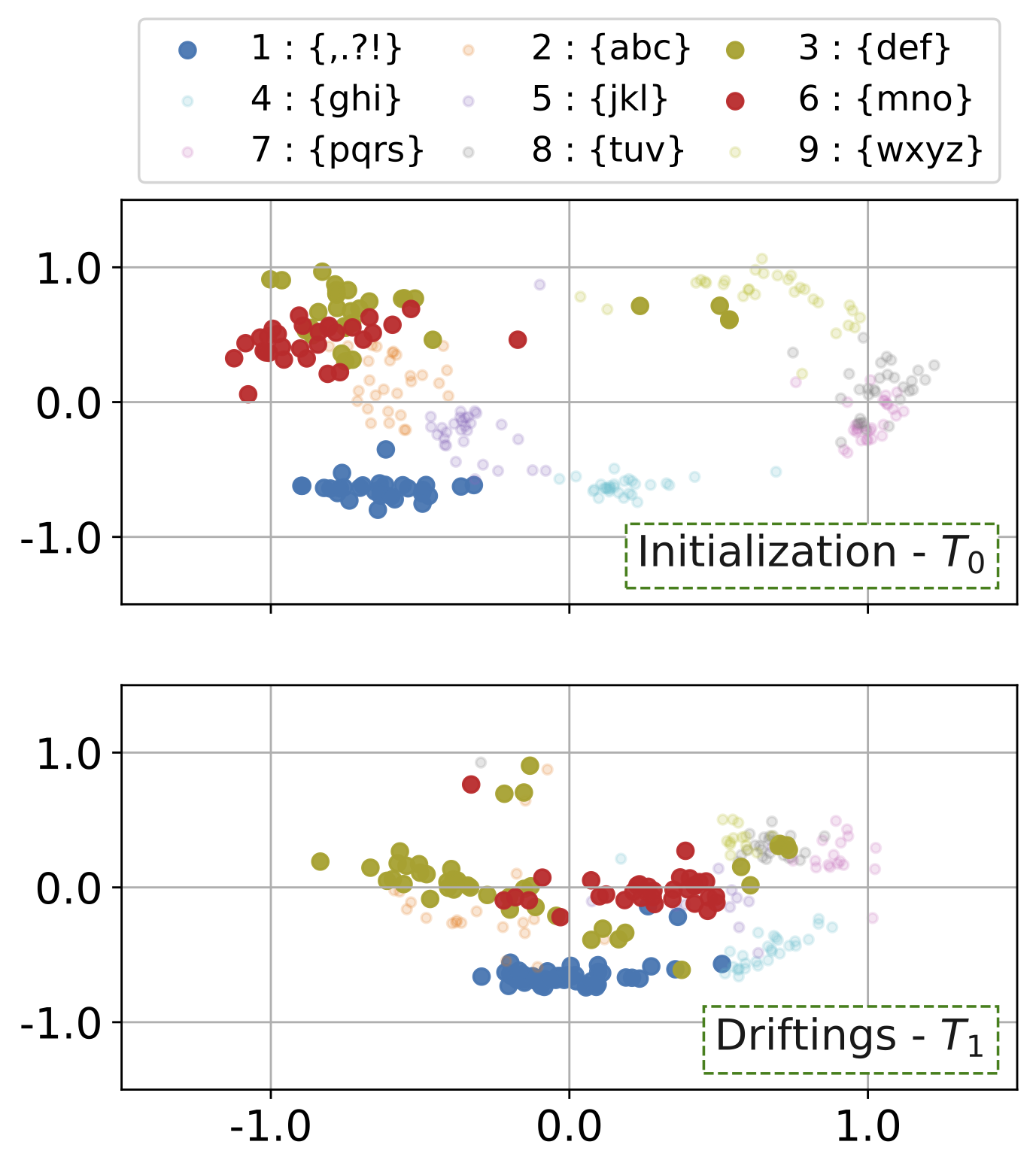}
        \caption*{\small (a) TAPRINT.}
    \end{minipage}
    \begin{minipage}[t]{0.4\linewidth}
        \centering 
        \includegraphics[width=0.95\linewidth]{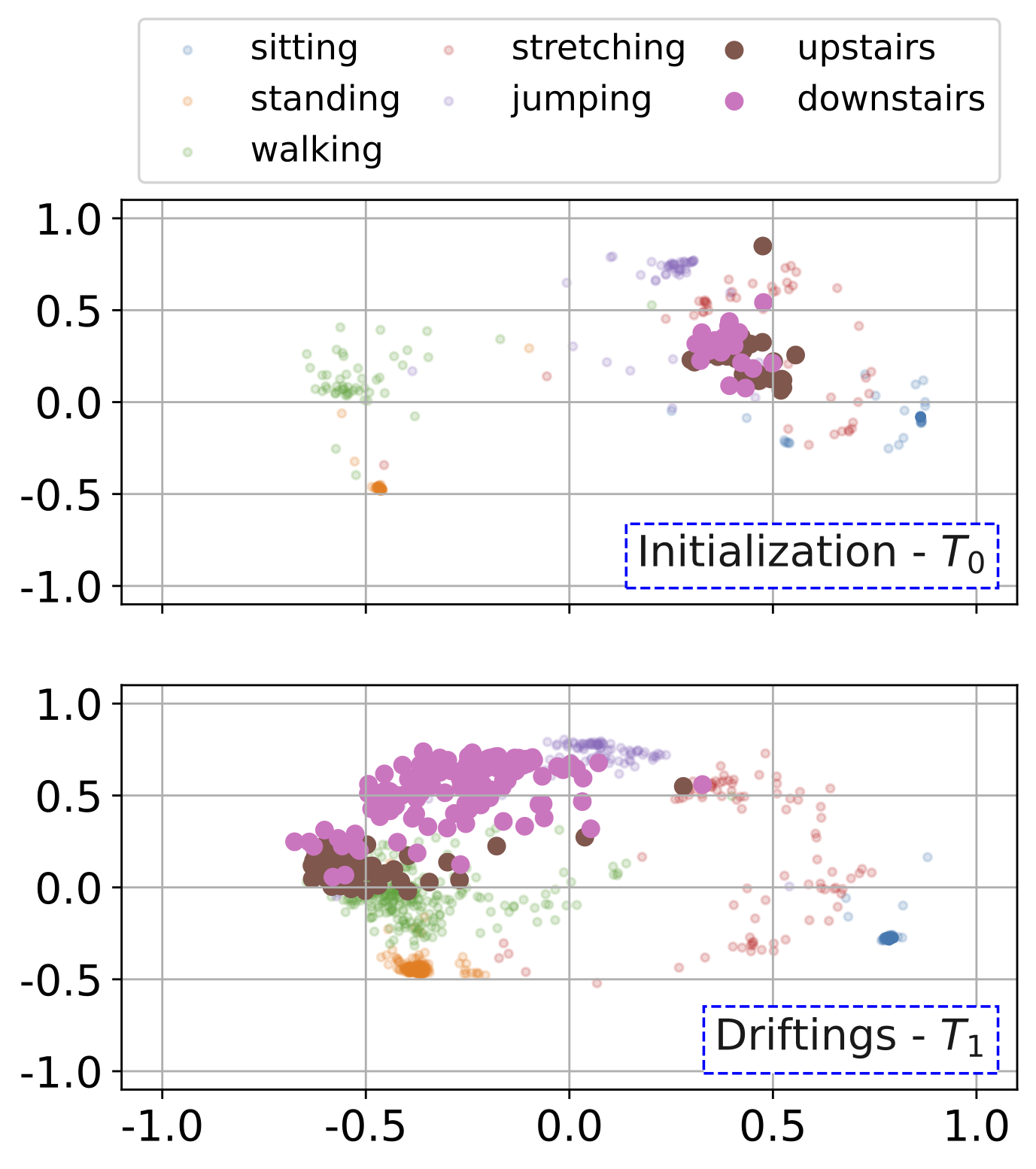}
        \caption*{\small (b) WHAR.}
    \end{minipage}
    
    \caption{The drfitings at time $T_1$ remains critical even with \underline{augmented} features (by MetaSense\cite{gong_metasense_2019}).}
    \vspace{-2em}
    \label{fig:motivate_2}
\end{figure}

\textit{Cache-based Selective Assimilation} is designed to address the uneven contribution of instances during continuous deployment.
It is motivated by a simple but robust observation:
across different usage stages, instances sharing the same label remain clustered in the feature space, 
while instances of different labels remain distinguishable—even when features are either manually engineered or learned by deep models (Fig.~\ref{fig:motivate_1} and~\ref{fig:motivate_2}).

\subsubsection{Cache-like Instance-driven Structure via Refined IWAL}
Motivated by this instance-level structure,
CODA adopts an instance based model (IB-k), realizing adaptation through a cache-like memory of representative instances rather than parameter retraining.

We consider sensor data as a continuous stream.
Let $X$ denote the input space and $Y$ the output space.
At each time step $t$, the agent observes an instance $x_t \in X$ together with its feedback $\hat{y}_t \in Y$.
CODA assigns each instance an importance probability $p_t$, computed under the current hypothesis space
$H_t = \{h : X \rightarrow Z\}$ using a customized loss function
$\ell : Z \times Y \rightarrow [0, \infty)$.
Following the importance-weighted active learning (IWAL) setting\cite{beygelzimer_importance_2009},
the loss is normalized to the range $[0,1]$ by assuming a bounded output space $Z$.
At time step $t$, CODA maintains an instance cache:
\begin{equation}
S_t = \{(x_i, y_i, w_i, Q_i, t_i)\}_{i=1}^{|S_t|},
\end{equation}
which serves both as the prediction basis and the adaptation interface.

\begin{figure}[t]
  \centering
    \includegraphics[width=.7\linewidth]{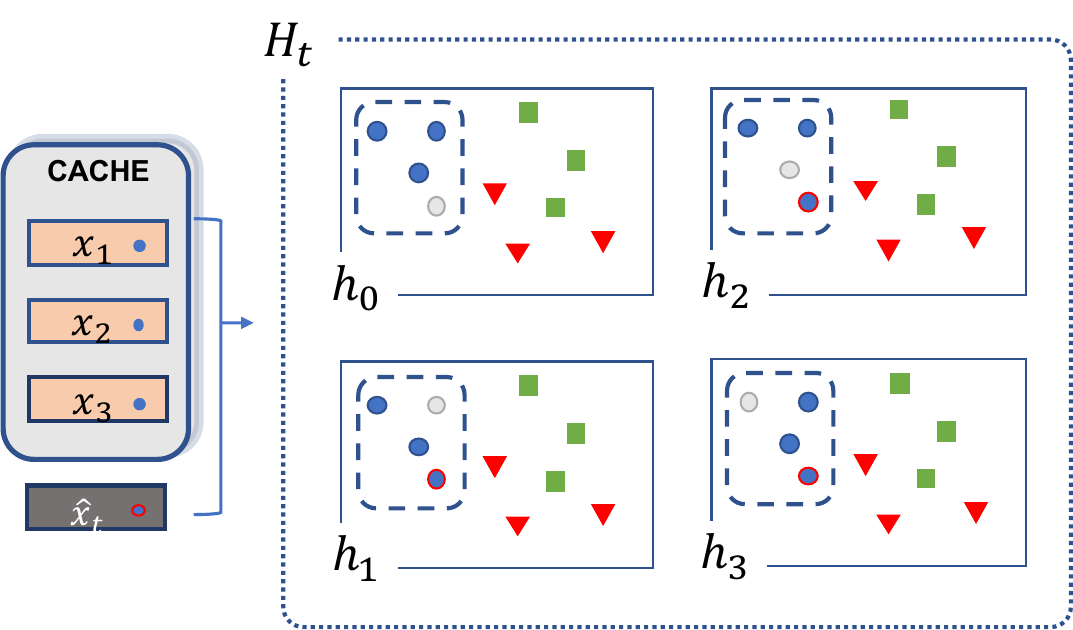}
  \caption{Hypothesis space for instance-based model (N = 3) }
  \label{fig:hs}
  \vspace{-2em}
\end{figure}

However, traditional IWAL assumes access to an oracle that can be actively queried for labels, an assumption that rarely holds in mobile sensing systems where supervision is sparse, delayed, and passively generated through user interaction.
To bridge this gap,
CODA fundamentally refines IWAL (Refined IWAL) to operate under passive supervision and instance-based prediction.
Instead of actively querying labels and retraining model parameters, Refined IWAL incorporates feedback $\hat{y}_t$ only when it naturally occurs and realizes adaptation by selectively updating cache contents. 


Algorithm~\ref{alg:IWAL} summarizes the overall adaptation procedure.
Upon receiving $(x_t, \hat{y}_t)$, CODA invokes the \textit{Rejection-Threshold} subroutine (Algorithm~\ref{alg:Rejection}), which evaluates the instance against the historical cache $S_{t-1}$.
As illustrated in Fig.~\ref{fig:hs}, for a fixed cache size $N$ per class, the hypothesis space $H_t = \{h_i\}_{i=1}^{N}$ is constructed by considering different combinations of cached instances within the same class, while instances from other classes remain unchanged.
The original hypothesis $h_{0}$ is excluded to encourage bolder evolution.
The \textit{Rejection-Threshold} subroutine returns a pruned hypothesis space $H_t$ together with an importance probability $p_t$, indicating the potential utility of the instance under the current conditions.

Only instances sampled according to $p_t$ are assimilated into the cache,
with importance weight $w_t = 1/p_t$.
The adapted hypothesis $h_t$ is then obtained by minimizing the Weighted Empirical Risk Minimization (Weighted ERM):
\begin{equation}
\mathcal{L}_t(h, S_t)
=
\frac{1}{|S_t|}
\sum_{(x,y,w,Q)\in S_t}
Q \cdot w \cdot \ell\big(h(x), y\big),
\label{func:expLoss}
\end{equation}
where $Q \sim \text{Bernoulli}(p_t)$ controls instance inclusion.

Notably, CODA does not rely on continuous or dense feedback.
In the absence of feedback, the system continues to operate using cached instances and importance estimation with the consistent prediction, while feedback, when available, acts as a corrective signal to refine future cache updates.

\subsubsection{Clustering Loss with Bounded Kernel}

The refined IWAL mechanism introduced above determines \emph{whether} an incoming instance should be assimilated into the cache by estimating its importance.
To quantify this importance, CODA designs a customized clustering loss with a bounded kernel for instance-level hypothesis evaluation. 

The design is motivated by a simple but critical observation in human activity recognition: 
\textbf{instances sharing the same activity label (e.g., walking) should exhibit high mutual similarity and form compact clusters in the representation space, while remaining well separated from instances of other activities.}
Deviation from this structure indicates ambiguity or potential misclassification and thus provides a meaningful signal for instance evaluation.

To operationalize this intuition, CODA introduces a soft clustering loss:
\begin{equation}
\ell_c(h(x), y)
=
\max
\left\{
0,
\hat{E}\big[d(\mathbf{X}_y, x)\big]
-
\min d(\overline{\mathbf{X}}_y, x)
\right\},
\label{func:clusLoss}
\end{equation}
where $\mathbf{X}_y$ denotes cached instances with label $y$, and $\overline{\mathbf{X}}_y$ denotes all remaining cached instances.
Intuitively, the loss penalizes instances whose average distance to the same-class cluster exceeds their minimum distance to other classes, indicating structural inconsistency in the cache.

To ensure compatibility with the theoretical guarantees of IWAL, the loss must be bounded.
CODA satisfies this requirement by constructing the distance function using a cosine-normalized kernel:
\begin{equation}
d(u, v)
=
\sqrt{
2\left(
1 -
\frac{\mathcal{K}(u,v)}
{\sqrt{\mathcal{K}(u,u)\mathcal{K}(v,v)}}
\right)
},
\label{func:disKernel}
\end{equation}
where $\mathcal{K}$ is a positive-definite kernel.
This formulation guarantees bounded distances while remaining flexible with respect to the choice of similarity measure.

CODA supports multiple kernel instantiations to accommodate different deployment constraints.
A linear kernel enables efficient computation for latency-sensitive scenarios, while the Global Alignment Kernel (GAK)~\cite{cuturi_fast_nodate} captures temporal misalignment in multivariate sensor sequences.
By decoupling the loss formulation from the kernel choice, CODA balances theoretical validity, computational efficiency, and representational power in continuous mobile sensing deployments.

\subsection{Adaptive Temporal Retention Strategy}

\begin{figure}[h]
    
    \setlength{\abovecaptionskip}{0.1cm}
    
    \begin{minipage}[b]{.416\linewidth}
      \setlength{\abovecaptionskip}{-0.1cm}
      \flushleft
      \includegraphics[width=.99\linewidth]{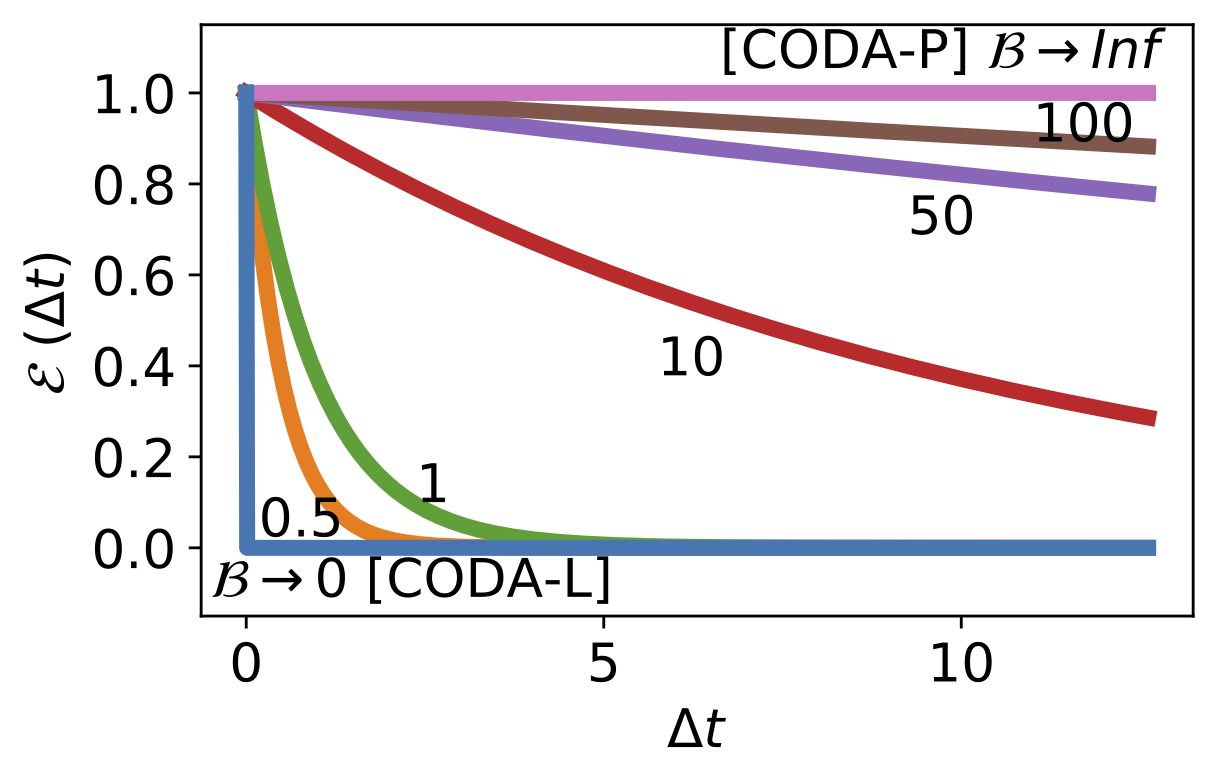}
      \caption*{\small{(a)}}
        
    \end{minipage}
    \begin{minipage}[b]{.554\linewidth}
      \setlength{\abovecaptionskip}{-0.1cm}
      \centering
      \includegraphics[width=.99\linewidth]{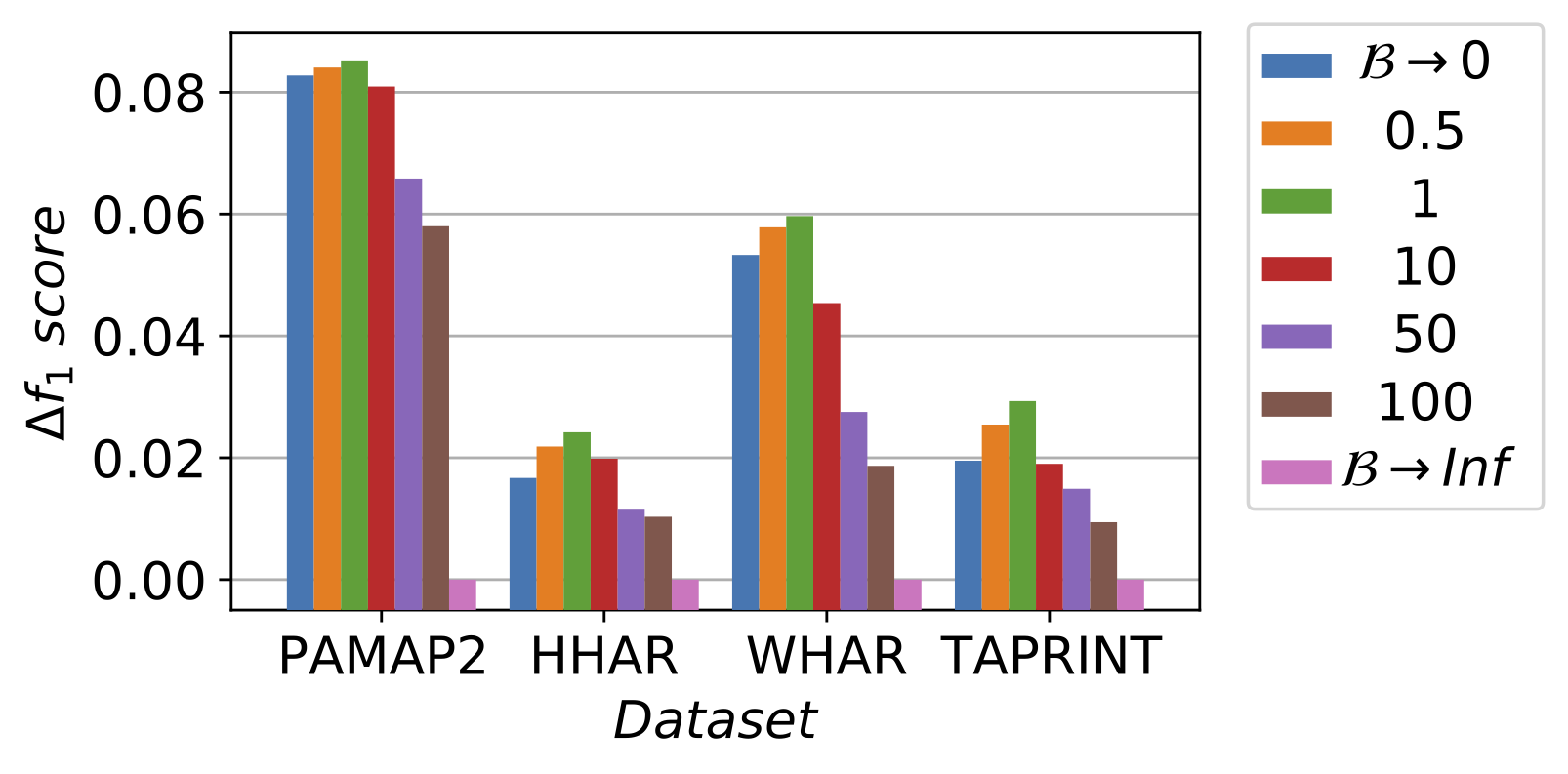}
      \caption*{\small{(b)}}

    \end{minipage}
    \caption{(a) Retentive functions varied by $\mathcal{B}$. (b) Performance improvement across $\mathcal{B}$ selection on different datasets.}
    \label{fig:retentiveWeightingResult}
\end{figure}

\subsubsection{Retentive Weighting Function}

Cache-based selective assimilation allows CODA to concentrate on informative instances, yet IWAL alone remains insufficient for long-term deployment under non-stationary sensing conditions.
In realistic deployment, feature distributions evolve continuously (Fig.~\ref{fig:motivate_1} and~\ref{fig:motivate_2}), causing instances collected in earlier phases to gradually lose relevance.
When such legacy instances are treated as equally valid, they may dominate the empirical risk and bias hypothesis evaluation, ultimately leading to misguided instance selection and degraded adaptation performance.

To explicitly address this issue, CODA incorporates temporal awareness into instance assessment via a retentive weighting function.
Rather than permanently assigning uniform influence to all cached instances, CODA modulates their contribution according to recency, enabling the adaptation process to continuously emphasize patterns that better reflect the current sensing context.
Formally, the time-aware loss for an instance $x$ at time $t$ is defined as:
\begin{equation}
  \ell_{t}(h(x), y;\, \Delta t_x)
  =
  \mathcal{E}_{t}(\Delta t_x)
  \cdot
  \ell_{c}\big(h(x), y\big),
  \label{func:jointLoss}
\end{equation}
where $\ell_{c}$ denotes the clustering loss defined in
Eq.~\eqref{func:clusLoss}, and $\Delta t_x = t - t_x$ represents the elapsed time
since instance $x$ was incorporated into the cache at time $t_x$.
The function $\mathcal{E}_t(\cdot)$ assigns a temporal retention weight that
controls how strongly historical instances contribute to hypothesis evaluation.

Inspired by the Ebbinghaus forgetting curve~\cite{ebbinghaus2013memory}, CODA
adopts a monotonically decreasing exponential form:
\begin{equation}
  \mathcal{E}_t(\Delta t_x)
  =
  \exp\!\left(-\frac{\Delta t_x}{\mathcal{B}}\right)
  \in (0, 1),
  \label{func:retenWeight}
\end{equation}
where $\mathcal{B}$ determines the effective temporal horizon of the agent.
Smaller values of $\mathcal{B}$ prioritize recent observations and facilitate
rapid adaptation to abrupt distributional shifts, whereas larger values preserve
longer-term context and stabilize learning under gradual drift.
In the limiting cases, $\mathcal{B} \rightarrow 0$ yields a memoryless evaluation
regime, while $\mathcal{B} \rightarrow \infty$ recovers standard importance-weighted
active learning without temporal decay.
Fig~\ref{fig:retentiveWeightingResult} illustrates this family of retention
functions as a function of relative timestep $\Delta t$.
Notably, the retentive weighting function operates solely at the evaluation
level: it modulates the influence of cached instances without altering the cache
composition itself.

\section{Evaluation} \label{section:Experiments}


This section evaluates CODA by addressing the following research questions:
\begin{itemize}[leftmargin=*]
    \item \textbf{RQ1}: Can one-off adaptation sustain performance under continuous domain drift, and does continuous adaptation offer clear advantages?
    \item \textbf{RQ2}: Can CODA reliably initialize and adapt without prior knowledge of the target user or device?
    \item \textbf{RQ3}: Is continuous adaptation effective and sustainable under long-term, sparse, and passive supervision?
    \item \textbf{RQ4}: Can CODA operate in real time on resource-constrained devices?
\end{itemize}

\newcolumntype{L}[1]{>{\raggedright\arraybackslash}p{#1}}
\newcolumntype{C}[1]{>{\centering\arraybackslash}p{#1}}
\newcolumntype{R}[1]{>{\raggedleft\arraybackslash}p{#1}}
\newcommand\crossRows[1][3]{\multirow[c]{#1}{*}}

\begin{table}[b]
\centering
  \setlength{\abovecaptionskip}{0.1cm}
  \setlength{\belowcaptionskip}{-0.1cm}
  \caption{Summary of datasets}
  \label{tab:datasetCfg_variance}
  \renewcommand\arraystretch{1.05}
  \scalebox{0.65}{
  \begin{tabular}{l l c l c c c}
    \toprule
    \textbf{Task} & \textbf{Dataset} & \textbf{\#Subjects} & \textbf{Device(s)}
    & \multicolumn{3}{c}{\textbf{Sample Rate (Hz)}} \\
    \cmidrule(lr){5-7}
    & & & & \multicolumn{2}{c}{\textbf{ACCE.}} & \textbf{GYRO.} \\
    \midrule

    \multirow{7}{*}{\textbf{ADLs}}
    & PAMAP2~\cite{reiss_creating_2012} & 5 & -- & 100 & 100 & 100 \\

    \cmidrule(lr){2-7}
    & \multirow{3}{*}{HHAR~\cite{stisen_smart_2015}} & \multirow{3}{*}{8}
    & Nexus 4     & \multicolumn{2}{c}{200} & -- \\
    &  &  & Galaxy S3  & \multicolumn{2}{c}{100} & -- \\
    &  &  & Samsung Old & \multicolumn{2}{c}{93}  & -- \\

    \cmidrule(lr){2-7}
    & \multirow{3}{*}{WHAR$^\dagger$} & \multirow{3}{*}{12}
    & LG & \multicolumn{2}{c}{200} & 200 \\
    &  &  & Huawei & \multicolumn{2}{c}{100} & 100 \\
    &  &  & TicWatch & \multicolumn{2}{c}{103} & 103 \\

    \midrule
    \textbf{GR} & TAPRINT$^\dagger$ & 9 & LG / Huawei / TicWatch
    & \multicolumn{3}{c}{Same as \textbf{WHAR}} \\
    \bottomrule
  \end{tabular}
  }
\end{table}

\subsection{Experiment Setup} \label{subsec:dataSets}

\subsubsection{Datasets Discriptions}
To evaluate CODA under diverse sensing conditions, we conduct experiments on two common tasks in human activity recognition, \textbf{A}ctivity for \textbf{D}aily \textbf{L}iving (as ADLs) recognition and \textbf{G}esture \textbf{R}ecognition (as GR).
Four human-centric sensing datasets covering multiple users, devices, and application scenarios, as showin in Table \ref{tab:datasetCfg_variance}.

\noindent\textbf{\textit{PAMAP2}}~\cite{reiss_creating_2012} is a activity recognition dataset with collected from multiple IMUs and treat measurements from all IMUs as a continuous multi-channel sensor stream with 27 channels (3 units × 3 sensors × 3 axes) and 5 subjects.

\noindent\textbf{\textit{HHAR}}~\cite{stisen_smart_2015} contains activity data collected from 8 users using heterogeneous smartphones and smartwatches.

\noindent\textbf{\textit{WHAR}$^\dagger$} is collected from 12 subjects using three smartwatch models for activity recognition.

\noindent\textbf{\textit{TAPRINT}$^\dagger$} is collected from 9 subjects on three smartwatch models, providing continuous gesture streams from re-implemented Taprint text-entry sessions~\cite{chen2019taprint}.

\begin{table}[tbp]
  \centering
  \setlength{\abovecaptionskip}{0.25cm}
  \setlength{\belowcaptionskip}{-0.15cm}
  \caption{Feature selection results}
  \label{tab:featSelection}
  \scalebox{.70}{
    \begin{tabular}{|c|c|C{14mm}|C{14mm}|C{14mm}|C{14mm}|}
    \bottomrule
    \multicolumn{1}{|c|}{\bfseries FEAT.} & {\bfseries METRIC} & {\bfseries PAMAP2} & {\bfseries HHAR} & {\bfseries WHAR} & {\bfseries TAPRINT} \\
    \hline
    ECDF & {EU} & \itshape \bfseries 0.8828 & \itshape \bfseries 0.8980 & \itshape \bfseries 0.8727 & 0.6856 \\
    \hline
    \multirow[c]{2}{*}{RAW} & EU & 0.8277 & 0.7451 & 0.7721 & 0.7657 \\
    \cline{2-2}
    & DTW & 0.8539 & 0.7014 & 0.6592 & \itshape \bfseries 0.7863 \\
    \toprule
    \end{tabular}
  }
\end{table}

\begin{table}[t]
  
  \setlength{\abovecaptionskip}{0.25cm}
  \setlength{\belowcaptionskip}{-0.15cm}
  \caption{Summary of experiments}
  \label{tab:datasetCfg}
  \renewcommand\arraystretch{1.01}
  \scalebox{.65}{
    \begin{tabular}{| c|c|c|c|c|C{13mm}|C{13mm}|C{13mm}|}
      \bottomrule
           \crossRows[2]{\bfseries{Dataset}} & \crossRows[2]{\bfseries{Class}}
           & \multicolumn{2}{c|}{\bfseries{Feature} }
           & \crossRows[2]{\bfseries{$\#$-Samples}}
           & \multicolumn{3}{c|}{\bfseries{Experiment(in \#-collections)}} \\
           \cline{3-4}\cline{6-8}
           &
           & \bfseries{$\#$-C} &  \bfseries{$\#$-W}
           & 
           & \bfseries{Online} & \bfseries{User} & \bfseries{Device} \\
      \hline
      \bfseries{PAMAP2\cite{reiss_creating_2012}} & 12 
          & 27 & \crossRows[3]{30} & 25877 
          & 5 & - & -  \\ 
          \cline{1-3} \cline{5-8}
          \bfseries{HHAR\cite{stisen_smart_2015}} & {6} 
          & {3} &  & {166673}
          & {96} & {56} & {16}  
          \\ 
          \cline{1-3} \cline{5-8}
          \bfseries{WHAR$^\dagger$} & {7}
          & {6} &  & {41032} 
          & {24} & {56} & {8} \\ 
      \hline
      \bfseries{TAPRINT}$^\dagger$ & 9 
      & 6 & 36 & 32580 & 27 & 54 & 54  \\ 
      \toprule

      \end{tabular}
  }
  \vspace{-1em}
\end{table}

\begin{table*}[htp]
  \centering
  \setlength{\abovecaptionskip}{0.25cm}
  \setlength{\belowcaptionskip}{-0.15cm}
  \caption{Detailed results of baselines with one-off adaptation.}
  \label{tab:exp_0_result}
  \scalebox{.605}{
    \begin{tabular}{|c|C{14mm}|C{14mm}|C{14mm}|C{15.5mm}|C{12.5mm}|C{14mm}|C{14mm}|C{14mm}|C{14mm}|C{14mm}|}
    \bottomrule
    {\bfseries DATASET} & {\bfseries PAMAP2} & \multicolumn{3}{c|}{\bfseries HHAR} & \multicolumn{3}{c|}{\bfseries WHAR} & \multicolumn{3}{c|}{\bfseries TAPRINT} \\
    \hline
    {\bfseries MODEL(s)} & {-} & {$nexus4^*$} & {s3} & {samsungold} & {$HW^*$} & {TW} & {LG} & {HW} & {TW} & {LG} \\
    \hline
    CrossValid. & \itshape \bfseries 0.8828 & \itshape \bfseries 0.9168 & \itshape \bfseries 0.8896 & \itshape \bfseries 0.8875 & \itshape \bfseries 0.8972 & \itshape \bfseries 0.8648 & \itshape \bfseries 0.8561 & \itshape \bfseries 0.7562 & \itshape \bfseries 0.7619 & \itshape \bfseries 0.8410 \\
    5NN & 0.1169 & 0.8095 & 0.7889 & 0.6757 & 0.5173 & 0.5689 & 0.5187 & 0.5197 & 0.4408 & 0.5717 \\
    MetaSense & 0.6327 & 0.8565 & 0.8594 & 0.7313 & 0.7790 & 0.7850 & 0.7921 & 0.6499 & 0.5407 & 0.6821 \\
    \toprule
    \end{tabular}
  }
  \vspace{-1em}
\end{table*}

\subsubsection{Baseline}
Given that different experiments target different aspects of continuous adaptation, we employ distinct baselines to provide meaningful and fair comparisons. 

\begin{itemize}[leftmargin=*]
    \item \textbf{CrossValid}:
A 10-fold cross-validation using a nearest-neighbor classifier on the entire dataset. This setting assumes full access to all deployment data and thus serves as the best achievable performance without online constraints.

\item \textbf{5NN}:
A standard KNN model represents the simplest practical baseline and reflects the performance of naive instance-based adaptation. The K is empirically set to 5.

\item \textbf{MetaSense}:
A meta-learning-based domain adaptation method for HAR. It leverages neural networks trained to adapt quickly to new domains and represents parameter-based adaptation under one-off or episodic settings.

\item \textbf{RND}:
When the cache reaches capacity, a previously stored instance is randomly replaced. Under sufficient feedback, this strategy reflects the minimal benefit that online feedback alone can provide without principled instance selection.

\item \textbf{TH}:
A time-based replacement strategy similar to least-recently-used (LRU), where older instances are preferentially removed. This baseline captures adaptation driven purely by recency, without considering instance utility.

\item \textbf{CODA-P}:
An ablated version of CODA that directly follows classical importance-weighted active learning. Instance importance is computed using all historical data, without temporal decay or retentive weighting.

\item \textbf{CODA-L}:
A loss-driven strategy that updates the cache based solely on instantaneous loss, replacing the most dissimilar instance at each step. It ignores historical contribution and temporal relevance.

\end{itemize}
In addition, we report two reference settings to contextualize performance bounds:
\noindent\textbf{LB (Lower Bound)}
A zero-feedback setting where the system uses its own predictions as feedback, representing the most pessimistic supervision scenario.
\noindent\textbf{UB (Upper Bound)}
A fruitful-feedback setting that assumes access to ground-truth labels at all times, providing an optimistic upper bound on adaptation performance.

 \subsubsection{Implementation}


Two feature extraction pipelines are considered to support instance-based modeling.
We extract ECDF features with 30 bins to obtain length-invariant representations, and alternatively construct RAW features by resampling each segment to a fixed sampling rate (25\,Hz for ADLs~\cite{stisen_smart_2015}, 200\,Hz for GR~\cite{chen2019taprint}).
RAW features are paired with either Euclidean distance (EU) or Dynamic Time Warping (DTW) to account for temporal misalignment.
Dataset-specific feature configurations are selected based on cross-validation performance (Table~\ref{tab:featSelection}), where ADLs generally favor ECDF features, while TAPRINT benefits from RAW features with DTW.
For online evaluation, all datasets are temporally ordered by arrival time and split chronologically to simulate realistic deployment, where the system is initialized with limited data and continuously adapts as new instances arrive.

\subsubsection{Metric}

To evaluate the performance of the system, in this paper, we adopt the Macro $F_1$-score as the main metric:
$$
\text{Macro } F_1\text{-score} = \frac{1}{N}  \sum\limits_{i = 0}^{N} F_1\text{-score}_i
$$
where the subscript $i$ denotes the class in the dataset. 
The macro $F_1$-score mitigates the impact by imbalanced data, and therefore is selected as the major performance indicator.

\begin{figure}[hb]
\vspace{-1em}
  \setlength{\belowcaptionskip}{-0.25cm}

  \setlength{\abovecaptionskip}{0.0625cm}

  \centering
  \includegraphics[width=.6\linewidth]{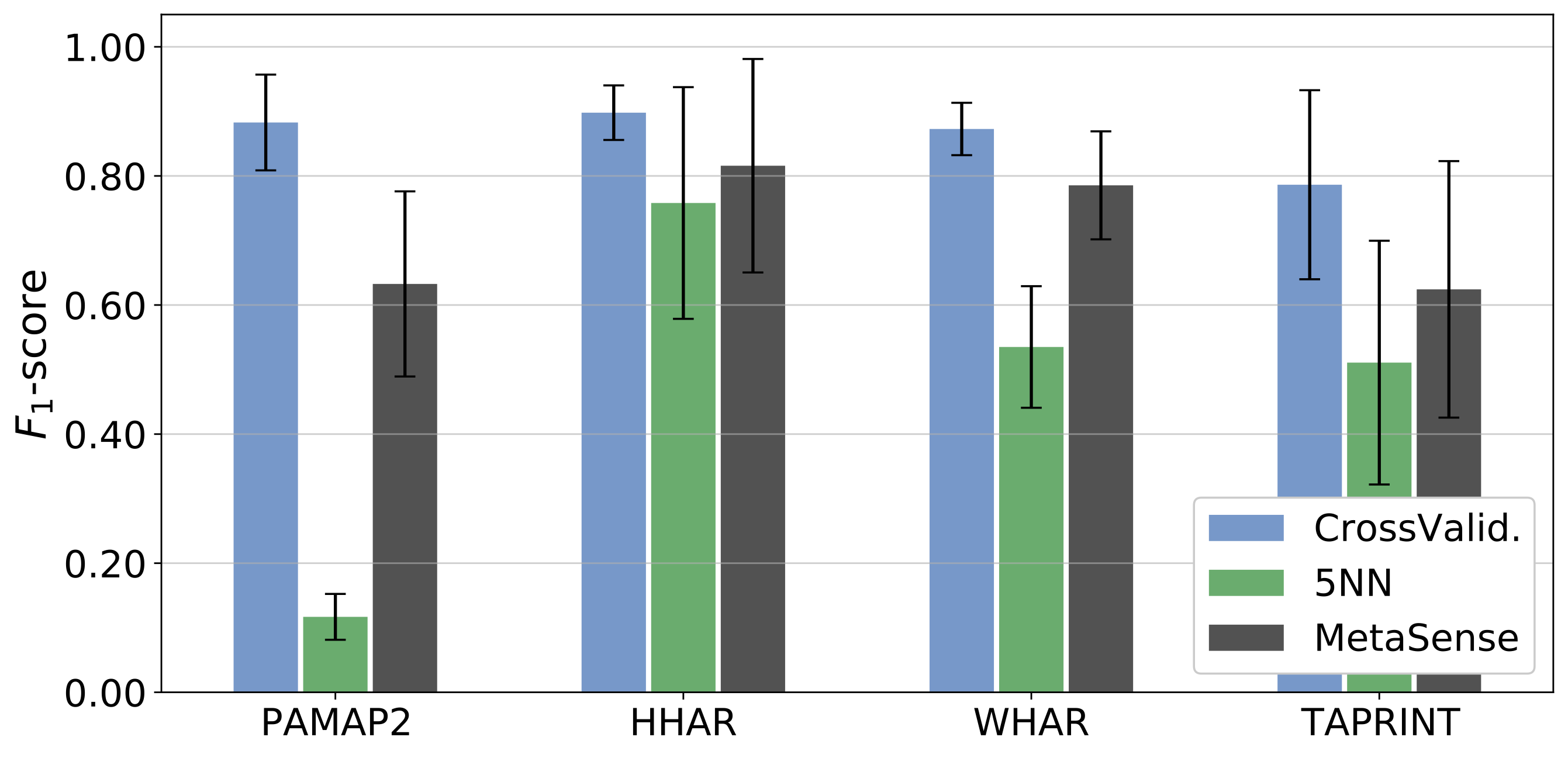} 
  \caption{Deployment with one-off domain adaptation}
  \label{fig:exp_0_motivation}

   

      
\end{figure}

\begin{figure}[tbp]
  \centering
  \vspace{-1em}
  \setlength{\abovecaptionskip}{0.1cm}
  \setlength{\belowcaptionskip}{-0.1cm}
    \begin{minipage}[t]{.486\linewidth}
    
        \centering
        \includegraphics[width=.6\linewidth]{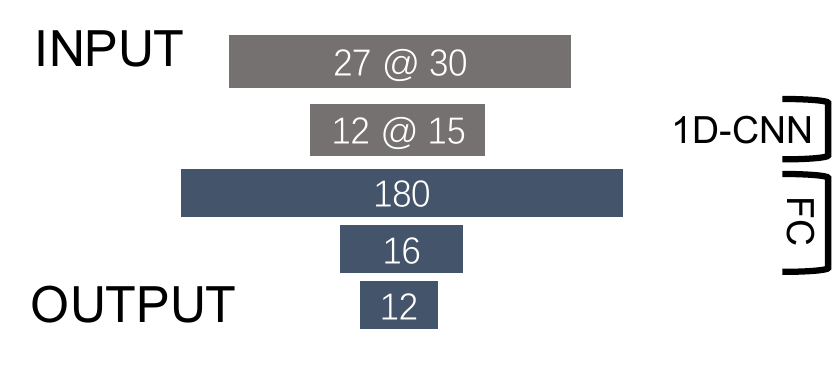} 
        \caption*{\small (a) PAMAP2.}
        
    \end{minipage}
    \begin{minipage}[t]{.396\linewidth}
     
        \centering
        \includegraphics[width=.6\linewidth]{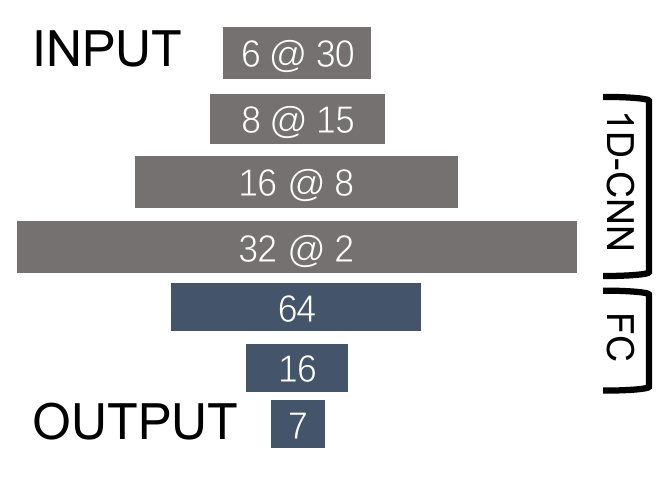} 
        \caption*{\small (c) WHAR.}
        
    \end{minipage}
    \begin{minipage}[t]{.486\linewidth}
    
        \centering
        \includegraphics[width=.6\linewidth]{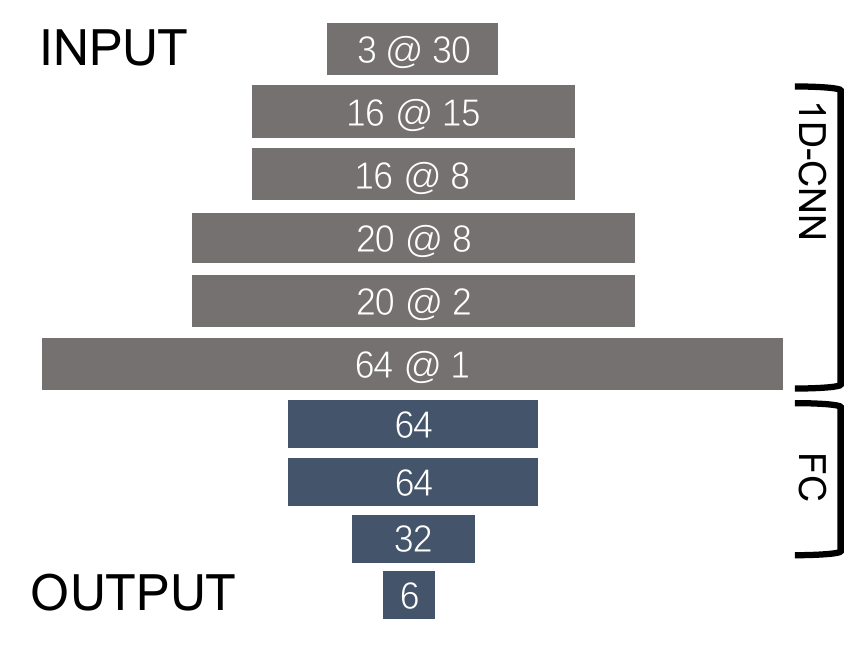} 
        \caption*{\small (b) HHAR.}
        
    \end{minipage}
    \begin{minipage}[t]{.396\linewidth}
     
        \centering
        \includegraphics[width=.6\linewidth]{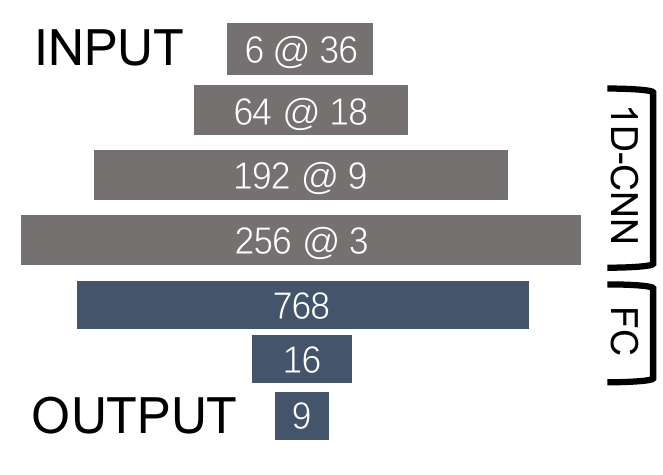} 
        \caption*{\small (d) TAPRINT.}
        
    \end{minipage}
    
  \setlength{\belowcaptionskip}{-0.25cm}
  \caption{Illustrations of structures. The output of 1D-CNN is flatterned as the augmented feature.}
  \label{fig:app_metaParams}
\end{figure}

\subsection{RQ1: One-off vs. Continuous Adaptation} \label{subsec:pilotStudy}

We study whether one-off adaptation can sustain performance under continuous domain drift, and whether continuous adaptation provides tangible benefits in long-term mobile sensing deployments. 
To this end, we compare a conventional one-off adaptation setting with sustained online adaptation under non-stationary conditions and reveals three insights:

\begin{itemize}[leftmargin=*]

\item \textbf{One-off adaptation cannot close the gap under continuous drift.} 
We first evaluate the one-off adaptation setting, where the model is adapted once using a limited amount of target data and then deployed without further updates.
This reflects a common assumption in prior work that a single adaptation stage suffices to handle deployment-time shifts.
To approximate the strongest one-off baseline, we train MetaSense using tasks derived from individual-condition datasets and select the best-performing configuration per dataset (Fig.~\ref{fig:app_metaParams}). 
As shown in Fig.~\ref{fig:exp_0_motivation} and Table~\ref{tab:exp_0_result}, the simple one-off baseline (5NN) fails to reach the performance of MetaSense~\cite{gong_metasense_2019}. 
However, even with MetaSense, one-off adaptation using limited target data remains substantially below the retrospective upper bound (CrossValid). 
This persistent gap across datasets indicates that a single adaptation step is insufficient to compensate for domain mismatch when deployment conditions evolve over time.

\item \textbf{Continuous adaptation sustains performance under non-stationary deployment.}
We evaluate CODA under a long-term deployment setting, where the system is initialized once and then continuously exposed to a data stream from multiple users in sequence, inducing sustained and non-stationary domain drift.
Under a fruitful feedback condition, the model updates online during operation and its accuracy is tracked over time.
As summarized in Table~\ref{tab:exp_5_longterm}, all online methods outperform the static 5NN baseline, confirming the necessity of adaptation.
Importantly, CODA consistently maintains a clear advantage over RND, demonstrating robust long-term adaptation rather than short-term overfitting.

\item \textbf{Effective continuous adaptation requires principled memory management.}
Further comparison among CODA variants reveals that long-term gains depend not only on receiving feedback, but on how historical information is retained.
CODA-P, which passively accumulates past instances, degrades over time due to reliance on outdated data.
In contrast, CODA-L and CODA mitigate this effect by periodically pruning cached instances based on temporal relevance.
These results show that continuous adaptation benefits arise from selectively assimilating informative samples while actively controlling memory staleness.
Unlike one-off adaptation, CODA enables the system to absorb new information over time and counteract accumulated drift, leading to sustained or even improving performance.

\end{itemize}

\begin{table}[t]

  \centering
  \setlength{\abovecaptionskip}{0.05cm}
  \caption{Long-term performance (in Accuracy) with \textbf{UB}.} 
  \label{tab:exp_5_longterm}

  \scalebox{.6}{
    \begin{tabular}{|l|cccccc|c|}
      \bottomrule
      {} & \multicolumn{2}{c}{LG} & \multicolumn{2}{c}{HW} & \multicolumn{2}{c|}{TW} & \crossRows[2]{AVG} \\
      {} & {WHAR} & {TAPRINT} & {WHAR} & {TAPRINT} & {WHAR} & {TAPRINT} &  \\
      \hline
      5NN & \color{gray} \underline 0.3497 & \color{gray} \underline 0.3076 & \color{gray} \underline 0.3642 & \color{gray} \underline 0.2634 & \color{gray} \underline 0.3650 & \color{gray} \underline 0.3559 & \color{gray} \underline 0.3343 \\
      \hline
      $\rhd$ RND & 0.8304 & 0.7888 & 0.8625 & 0.7424 & 0.8019 & 0.6918 & 0.7863 \\
      \hline
      TH & \color{gray} \underline 0.8133 & \color{black} 0.7972 & \color{gray} \underline 0.8328 & \color{black} 0.7470 & \color{gray} \underline 0.7875 & \bfseries \color{black} 0.7071 & \color{gray} \underline 0.7808 \\
      CODA-P & \color{gray} \underline 0.5575 & \color{gray} \underline 0.7103 & \color{gray} \underline 0.5678 & \color{gray} \underline 0.6395 & \color{gray} \underline 0.4894 & \color{gray} \underline 0.5721 & \color{gray} \underline 0.5894 \\
      CODA-L & \color{gray} \underline 0.8219 & \color{gray} \underline 0.7844 & \color{gray} \underline 0.8563 & \color{black} 0.7459 & \color{black} 0.8133 & \color{black} 0.6946 & \color{gray} \underline 0.7860 \\
      CODA & \bfseries \color{black} 0.8460 & \bfseries \color{black} 0.7991 & \bfseries \color{black} 0.8687 & \bfseries \color{black} 0.7545 & \bfseries \color{black} 0.8197 & \color{black} 0.6949 & \bfseries \color{black} 0.7971 \\
    \toprule
    \end{tabular}

  }
\vspace{-1em}
\end{table}

\begin{figure}[h]
    \centering
    \includegraphics[width=.95\linewidth]{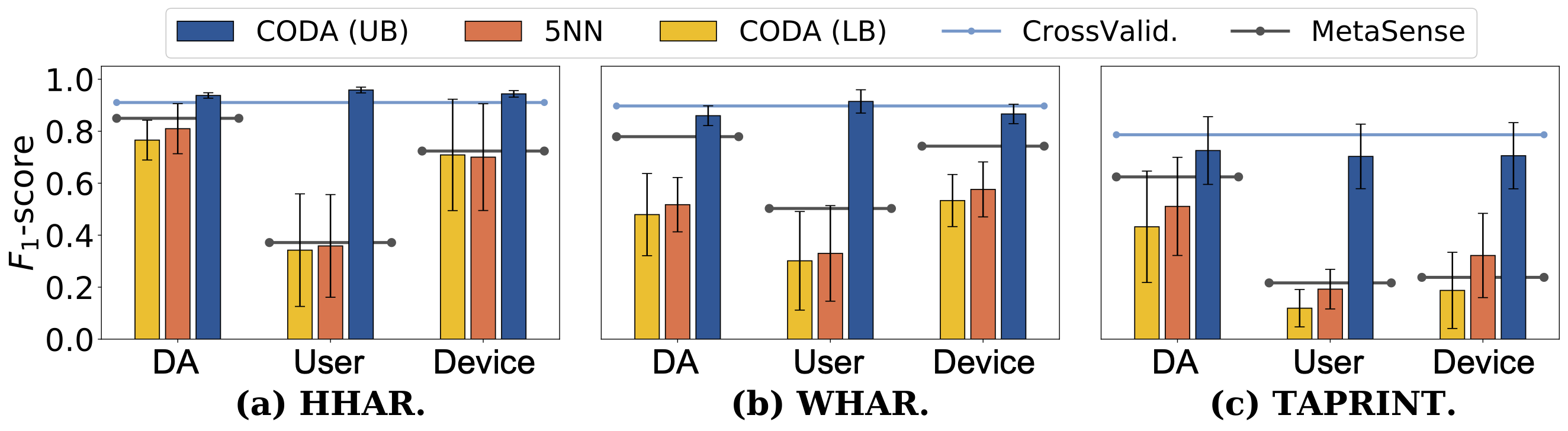}
  \caption{Target agnostic adaptation.}
    \label{fig:exp_3_VarySrcDomain}
    \vspace{-1em}
\end{figure}

\subsection{RQ2: Target Agnostic Adaptation} \label{subsec:TAD}
We define the \emph{target domain} as the data trace that requires adaptation.
Specifically, we construct two target-agnostic settings by initializing the
system with data that differ from the target domain either in terms of \textbf{user} or \textbf{device model}.
In the \textbf{User} setting, the system is initialized using data collected from a different user but the same device model.
In contrast, the \textbf{Device} setting initializes the system using data from the same user but a different device model.
\begin{itemize}[leftmargin=*]

\item \textbf{Setting (Table~\ref{tab:datasetCfg})}: For HHAR and WHAR (PAMAP2 is omitted since it involves only a single device model), we select collections from one out of three device models (marked with $^*$  in Table~\ref{tab:exp_0_result}) as the target domain, resulting in eight target collections per dataset.
User-agnostic evaluation is conducted by permuting these collections across
different users, yielding $8 \times 7$ adaptation traces for each dataset.
Device-agnostic evaluation initializes the system using data from different device models, producing 16 traces for HHAR and 8 traces for WHAR.
For TAPRINT, all three device models are treated as target domains, resulting in $3 \times 9$ collections.
To ensure balanced evaluation, the nine subjects are randomly partitioned into three groups, yielding 54 adaptation traces for each experimental condition.

\item \textbf{Result}: We compare the User-agnostic and Device-agnostic settings with a conventional
domain adaptation baseline (\textbf{DA}) that assumes partial prior knowledge of the target domain.
As shown in Fig.~\ref{fig:exp_3_VarySrcDomain}, variations across users generally lead to larger performance degradation than variations across devices, indicating that behavioral differences among users pose a greater challenge than sensor heterogeneity.
Despite this mismatch, CODA under the fruitful feedback (\textbf{UB}) condition exhibits consistent performance improvement across all experimental groups compared with other baseline.
This consistency demonstrates that CODA can adapt even when initialized with minimal and mismatched prior knowledge, thereby validating its target-agnostic adaptation capability.
\end{itemize}

\begin{figure}[t]
    \centering
    \setlength{\abovecaptionskip}{0.025cm}
    \setlength{\belowcaptionskip}{-0.15cm}
    \includegraphics[width=.95\linewidth]{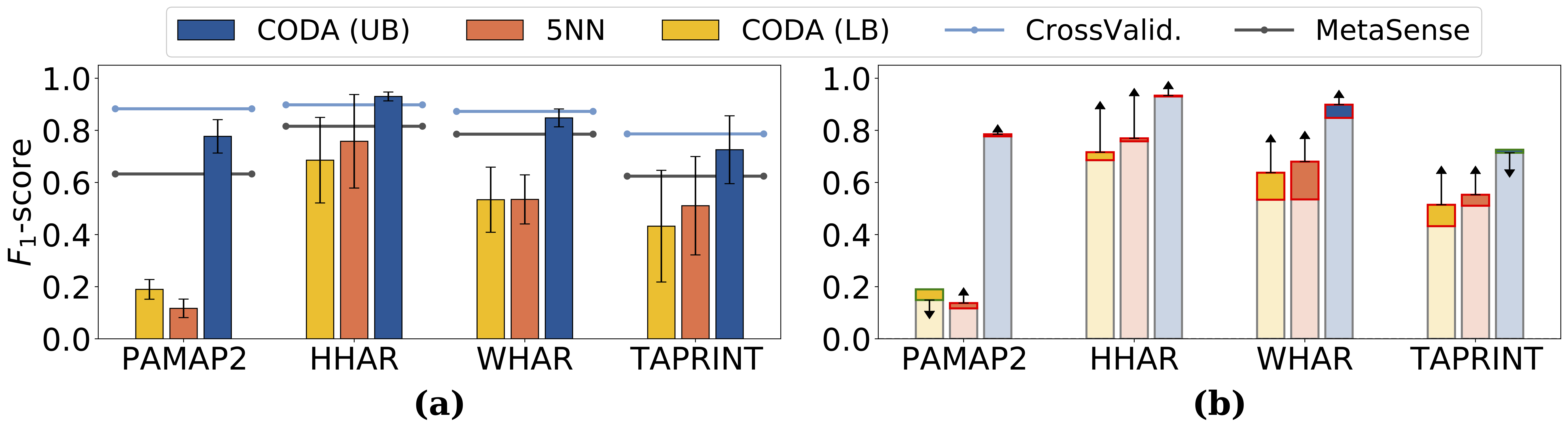}
  \caption{(a) Online adaptation results. (b) Augmented online adaptation results.}
  \vspace{-1em}
    \label{fig:exp_1add2_feedback}
\end{figure}

\begin{figure}[h]
    \centering
    \setlength{\abovecaptionskip}{0.025cm}
    \setlength{\belowcaptionskip}{-0.35cm}
    \includegraphics[width=.95\linewidth]{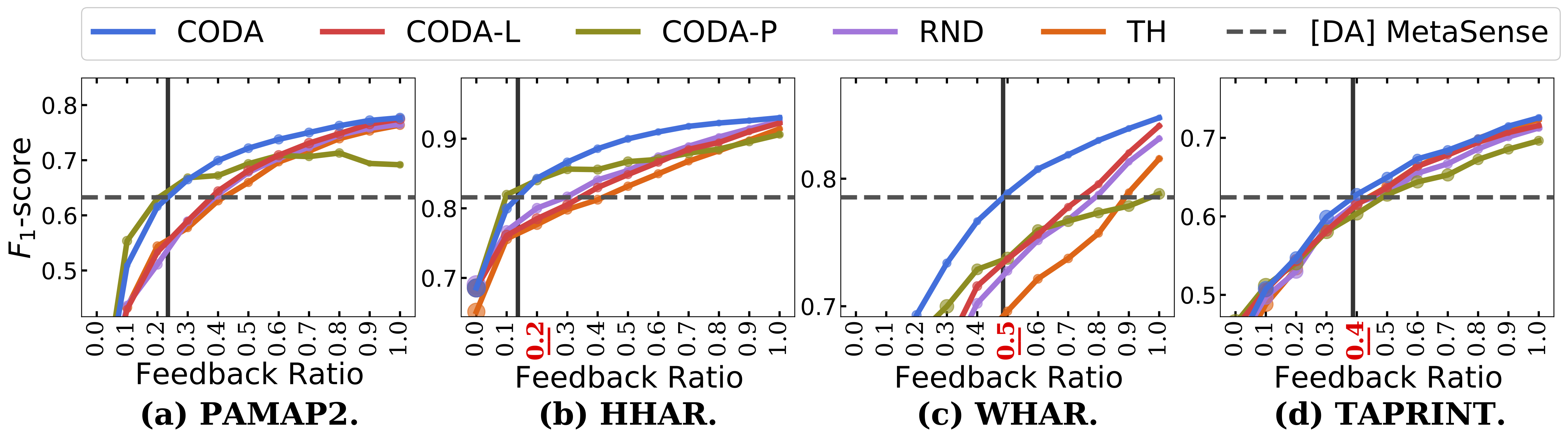}
  \caption{Practical online domain adaptation. The underlined ratios are Minimum Feedback Ratio. }
    \label{fig:exp_5_practical_setting_1}
    \vspace{-.8em}
\end{figure}

\begin{figure}[h]
    \centering
    \setlength{\abovecaptionskip}{0.025cm}
    \setlength{\belowcaptionskip}{-0.35cm}
    \includegraphics[width=.95\linewidth]{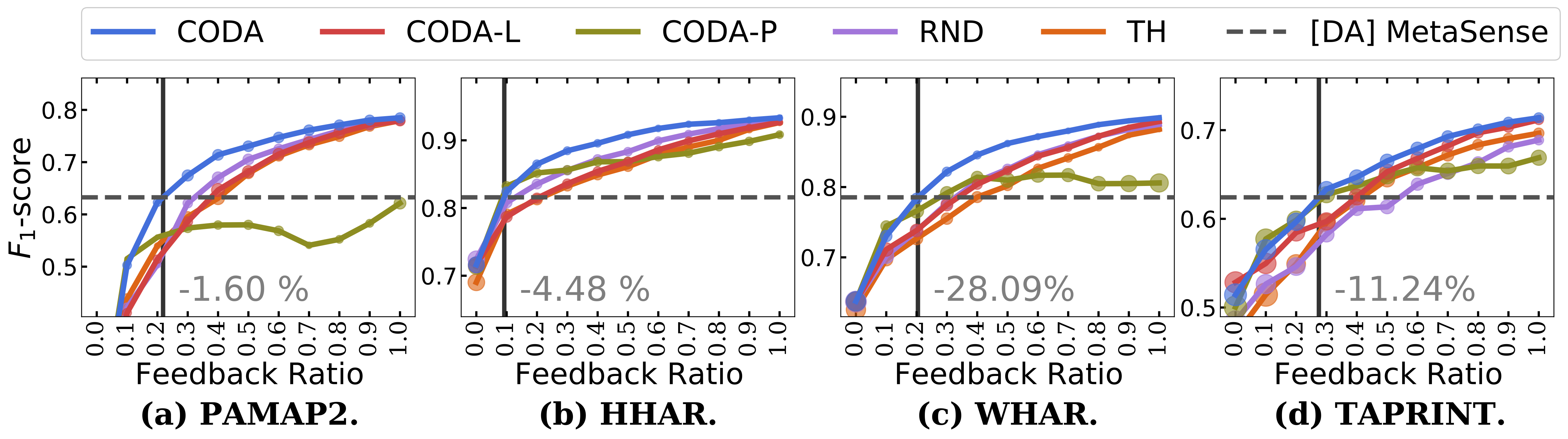}
  \caption{Practical online domain adaptation (augmented with MetaSense).
  The size of a marker only represents the relative value of standard deviation. }
    \label{fig:exp_5_practical_setting_2}
    \vspace{-0.5em}
\end{figure}

\subsection{RQ3: Impact of Feedback in Continuous Adaptation}

\begin{figure}[t]
    \centering
    \setlength{\abovecaptionskip}{0.025cm}
    \setlength{\belowcaptionskip}{-0.25cm}
    \includegraphics[width=.6\linewidth]{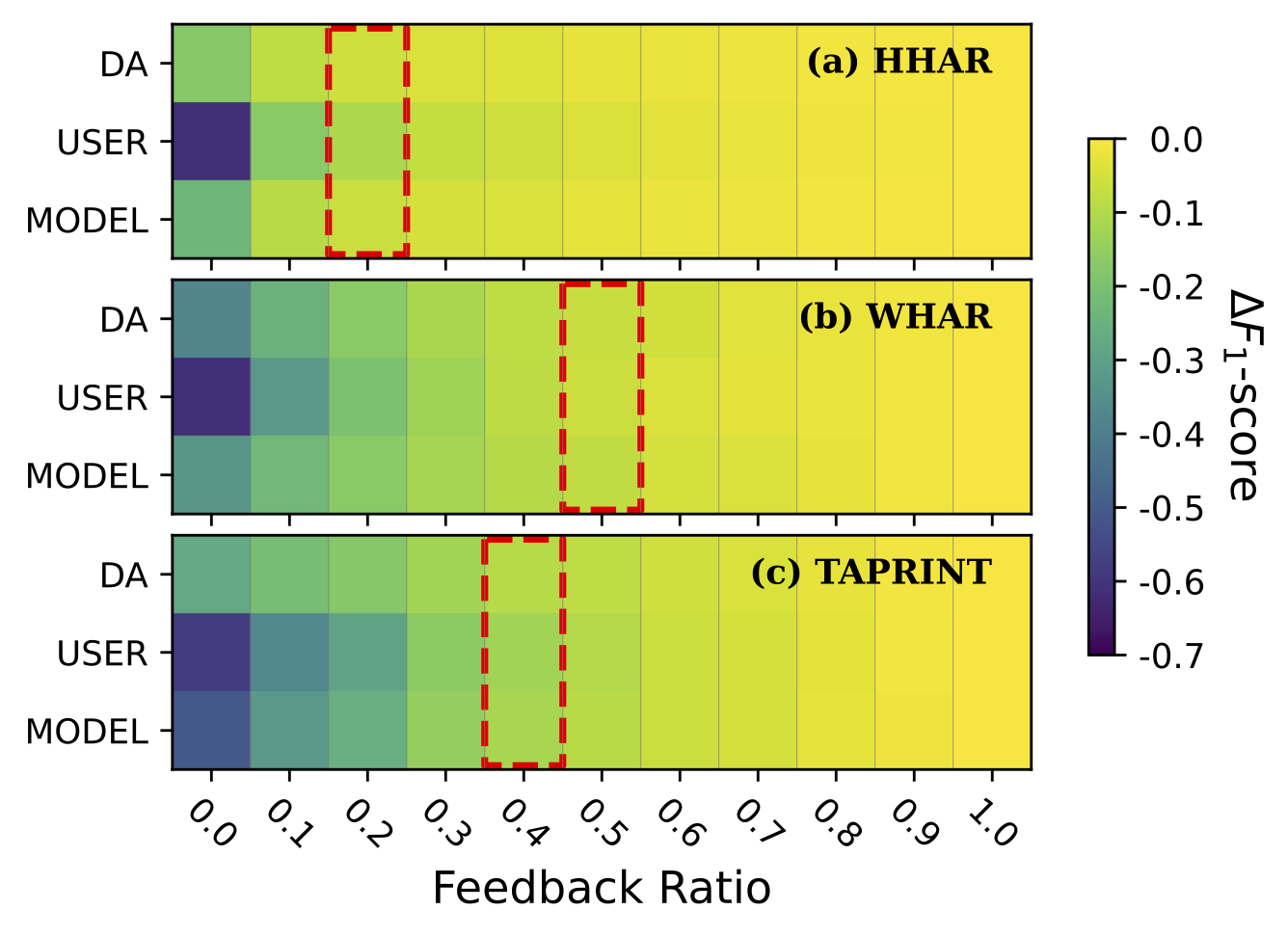}
  \caption{Practical adaptation on agnostic target. The variance by conditions diminishes as feedback ratio increases.}
    \label{fig:exp_5_practical_setting_3_varSrc}
    \vspace{-0.5em}
\end{figure}

As a continuously evolving framework during deployment, CODA is inherently influenced by the availability and quality of online feedback. We therefore examine the role of feedback from coarse to fine granularity, progressively analyzing its boundary effects, interaction with representation quality, and practical sustainability.
\begin{itemize}[leftmargin=*]

\item \textbf{Sensitive to feedback quality.}
We evaluate CODA under two extreme online conditions: \textbf{LB} and \textbf{UB}.
As shown in Fig.~\ref{fig:exp_1add2_feedback}(a), these settings lead to markedly different behaviors.
Under \textbf{LB}, performance on multiple datasets drops significantly below the 5NN baseline, indicating error accumulation caused by incorrect self-predictions.
In contrast, under \textbf{UB}, all online baselines consistently outperform 5NN in terms of macro-$F_1$, with some even surpassing the neural MetaSense baseline.
These results highlight the effectiveness of online adaptation and its sensitivity on feedback quality.

\item \textbf{Better representations reduce dependence on feedback.}
Owing to its modular design, CODA can directly integrate the feature extractor from MetaSense.
As shown in Fig.~\ref{fig:exp_1add2_feedback}(b), improved representations yield substantial gains under zero-feedback conditions, while providing only marginal improvements under fruitful feedback.
This asymmetry suggests that stronger representations primarily reduce reliance on feedback rather than amplify the benefits of dense supervision.
To examine whether this effect persists in realistic deployments, we further evaluate adaptation under \emph{limited} feedback.
As shown in Fig.~\ref{fig:exp_5_practical_setting_1} and Fig.~\ref{fig:exp_5_practical_setting_2}, CODA consistently outperforms competing baselines under partial feedback, benefiting from its IWAL-based instance selection.
We further define the \textbf{Minimum Feedback Ratio (MFR)} as the lowest feedback level at which continuous adaptation surpasses MetaSense.
After adopting MetaSense-trained feature extractor, the MFR is reduced by 2.94\% to 37.30\% across datasets, providing quantitative evidence that improved representations systematically lower the feedback required for effective adaptation.
These results indicate that CODA complements deep models by alleviating feedback demands rather than relying on dense supervision.

\item \textbf{Sufficient feedback enables target-agnostic adaptation.}
We examine target-agnostic adaptation under varying feedback ratios by measuring performance degradation relative to the ideal Macro-$F_1$ score.
As shown in Fig.~\ref{fig:exp_5_practical_setting_3_varSrc}, increasing feedback yields diminishing marginal returns.
Once the feedback ratio reaches approximately 0.7 across datasets, performance differences among user- and device-specific groups consistently shrink to within 1\%.
Beyond this point, additional feedback provides limited benefit, indicating that CODA has acquired sufficiently generalizable adaptation.

\end{itemize}

\begin{figure}[t]
    \centering
    \setlength{\abovecaptionskip}{0.025cm}
    \setlength{\belowcaptionskip}{-0.25cm}
    \includegraphics[width=.5\linewidth]{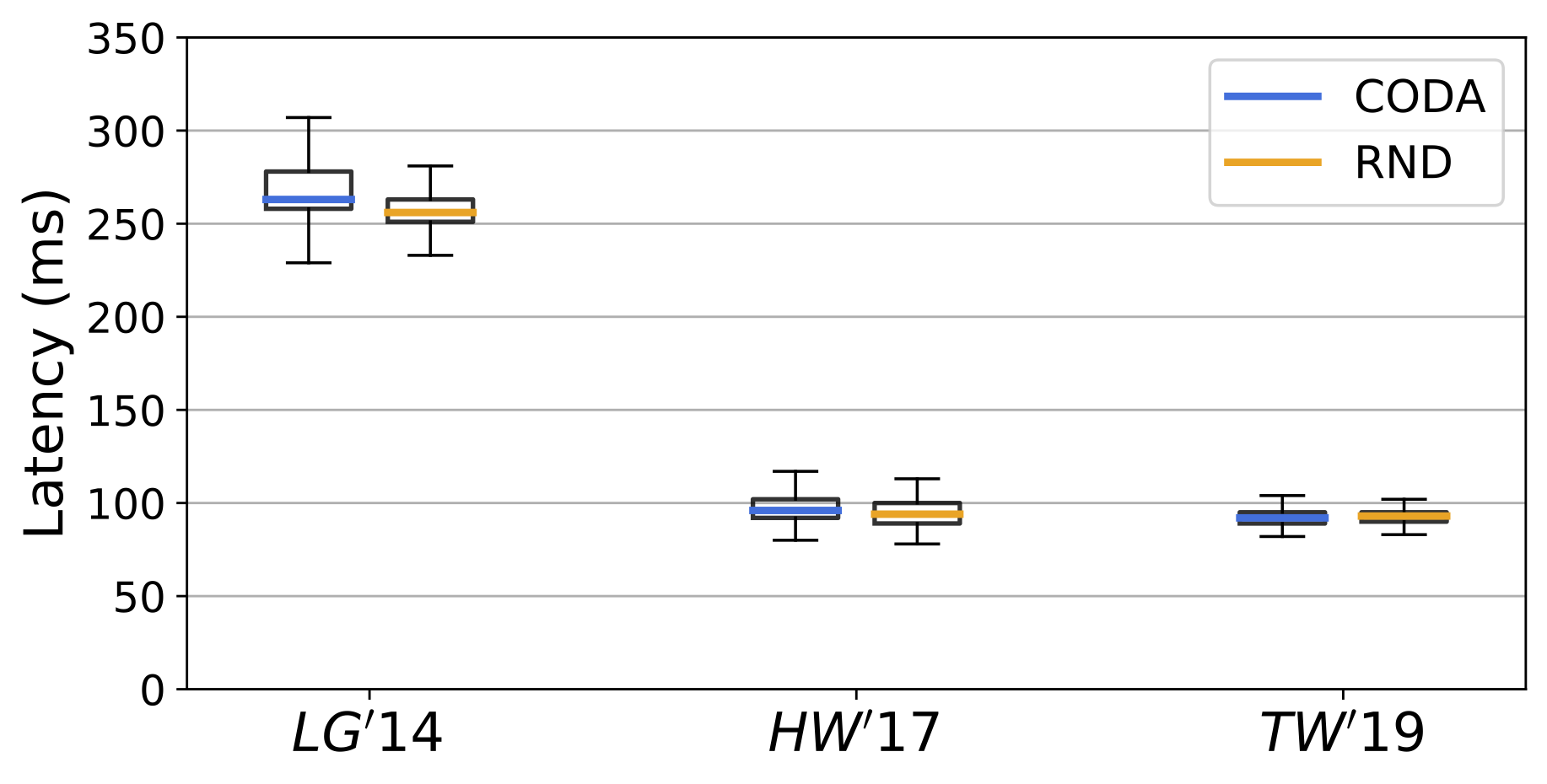}
  \caption{Comparison of latency (ms) by prediciton and adaptation (smartwatch  with the release date).}
    \label{fig:exp_5_practical_setting_4_impl}
    \vspace{-.7em}
\end{figure}


\subsection{RQ4: Latency and Real-world Use on Smartwatches} \label{subsec:impl}


We further investigate the practical feasibility of CODA through runtime latency in a real-world interactive application.

\begin{itemize}[leftmargin=*]

\item \textbf{Latency Overhead}: The end-to-end latency across different smartwatch brands is reported in Fig.~\ref{fig:exp_5_practical_setting_4_impl}.
Despite variations in hardware configurations, CODA is consistently compared against the RND to isolate the computational overhead introduced by adaptive memory management.
In particular, the median latency introduced by CODA increases by 7,ms on LG devices, 2,ms on Huawei (HW), and is reduced by 1,ms on TicWatch (TW).
These results indicate that the additional computational cost required to achieve stronger adaptation is modest and well within the practical constraints of on-device deployment.

\item \textbf{Real-world Application}: To further evaluate usability in realistic scenarios, we design an interactive gesture collection application following the Taprint protocol.
Specifically, we develop a gamified application, MAZE, which embeds gesture acquisition into an exploratory maze game using visual anchors, temporal pacing, and dynamic map selection.
For comparison, we also implement a non-gamified control application (CTRL) for gesture collection, as illustrated in Fig.~\ref{fig:exp_5_practical_setting_4_impl}.
After a 15-minute usage session, CODA demonstrates substantial performance improvements in both settings, achieving a relative gain of 15.4\% in CTRL and 19.6\% in MAZE.
These results suggest that CODA can effectively leverage user interactions, particularly when feedback is naturally integrated into engaging experiences, thereby alleviating the practical challenge of acquiring explicit user feedback in real-world deployments.
\end{itemize}

\section{Discussion: Limitations and Future Directions} 
Although CODA demonstrates strong performance, there remains room for improvement.
\begin{itemize}[leftmargin=*]

\item 
Its primary limitation lies in the reliance on user feedback, which is often imperfect and highly context-dependent in real-world deployments. 
Although our experiments systematically analyze the effects of feedback quality, quantity, and timing—and further validate feasibility through an interactive case study—designing unobtrusive feedback mechanisms during daily use remains an open challenge. 
Future systems may therefore shift from maximizing feedback frequency toward interaction designs that elicit high-value, low-burden feedback, such as opportunistic corrections, implicit cues, or task-driven confirmations.

\item The cache-based abstraction underlying CODA is not limited to the mobile sensing tasks studied in this work. 
The framework can be naturally extended to other continuously deployed systems, including personalized human–computer interaction, and context-aware IoT applications, where non-stationary data streams and delayed supervision are the norm. 
Exploring how CODA interacts with richer sensing modalities, multimodal fusion, and foundation-model-based representations~\cite{qiu2025towards} represents a promising direction for future research. 
We hope this work inspires the community to embrace continuous adaptation and to explore novel mobile sensing scenarios in which systems evolve alongside their users and environments.
\end{itemize}

\section{Conclusion}   
In this paper, we have presented CODA, a continuous online adaptation framework that adapts the mobile sensing system to uncertain conditions. 
The key innovation is to regard the online drifting conditions as the results from changes in data distribution.
Integrated with \textit{Cache-based Selective Assimilation}  and \textit{Adaptive Temporal Retention Strategy}, CODA achieves robust adaptation with trivial cache-like structure even without learnable parameters. 
The comprehensive experiments on four datasets (two of which are self-collected in this paper) demonstrates the feasibility and potential of the online adaptation.
As we propose to commit adaptation online, our following research would try to address the requirements of feedback by specifying the application design.
We hope that our study can inspire relevant research to contribute more novel scenarios with continuous online adaptation in mobile sensing.
\section*{Acknowledgment}
This research is supported in part by the Guangdong Provincial Key Lab of Integrated Communication, Sensing and Computation for Ubiquitous Internet of Things (No. 2023B1212010007), China NSFC Grant (No. 62472366, 62372307), the Project of DEGP (No. 2023KCXTD042, 2024GCZX003), Guangdong NSF (No. 2024A1515011691), “111 Center (No. D25008)”, Shenzhen Science and Technology Foundation (No. ZDSYS20190902092853047, JCYJ20230808105906014), Shenzhen Science and Technology Program (No. RCYX20231211090129039).







\bibliographystyle{ieeetr}
\bibliography{secon}

\vspace{12pt}

\end{document}